\documentclass[journal]{IEEEtran}
\ifCLASSINFOpdf
\else
\fi

\usepackage{amsmath}

\usepackage{algorithmic}

\usepackage{array}
\usepackage{verbatim}

\usepackage{graphicx}
\usepackage{subcaption}
\graphicspath{ {./images/} }

\usepackage{booktabs} % for better looking tables
\usepackage{multirow} % for multi-row cells
\usepackage{amssymb}
\usepackage{bbding}
\begin{document}

\title{Wi-Spike: A Low-power WiFi Human Multi-action Recognition Model with Spiking Neural Networks }

\author{Nengbo~Zhang$^{1}$,
       Yao~Ying$^{2}$,
       Lu~Wang$^{3}$,
       Kaishun~Wu$^{4}$,
       Jieming Ma$^{5}$,
       Fei~Luo$^{1}$,
       
              % <-this % stops a space
\thanks{$^{1}$Nengbo Zhang, Fei Luo (corresponding author) are with the School of Computing and Information Technology, Great Bay University, Dongguan, China.
e-mail:( bookerznb@gmail.com; luofei2018@outlook.com).}% <-this % stops a space
\thanks{$^{2}$Yao Ying is with the School of Computer Science, Guangdong University of Petrochemical Technology, Maoming, China. e-mail:( ying\_yao@foxmail.com).}
\thanks{$^{3}$Lu Wang is with the College of Computer Science and Software Engineering, Shenzhen University, Shenzhen, China. e-mail:( wanglu@szu.edu.cn).}% <-this % stops a space
\thanks{$^{4}$Kaishun Wu is with the Hong Kong University of Science and Technology (Guangzhou), Guangzhou, China. e-mail:( wuks@hkust-gz.edu.cn).}% <-this % stops a space
\thanks{$^{5}$Jieming Ma  is with the Department of Computer Science and Technology, Harbin Institute of Technology, Shenzhe, China. e-mail:( 24b951052@stu.hit.edu.cn).}% <-this % stops a space

\thanks{Manuscript received October 3, 2025; revised October 26, 2025.}}

% The paper headers
\markboth{IEEE Transactions on Mobile Computing, Submitted}%
{Shell \MakeLowercase{\textit{et al.}}: Bare Demo of IEEEtran.cls for IEEE Journals}

% make the title area
\maketitle

% As a general rule, do not put math, special symbols or citations
% in the abstract or keywords.
\begin{abstract}
WiFi-based human action recognition (HAR) has gained significant attention due to its non-intrusive and privacy-preserving nature. However, most existing WiFi sensing models predominantly focus on improving recognition accuracy, while issues of power consumption and energy efficiency remain insufficiently discussed. In this work, we present Wi-Spike, a bio-inspired spiking neural network (SNN) framework for efficient and accurate action recognition using WiFi channel state information (CSI) signals. Specifically, leveraging the event-driven and low-power characteristics of SNNs, Wi-Spike introduces spiking convolutional layers for spatio-temporal feature extraction and a novel temporal attention mechanism to enhance discriminative representation. The extracted features are subsequently encoded and classified through spiking fully connected layers and a voting layer. Comprehensive experiments on three benchmark datasets (NTU-Fi-HAR, NTU-Fi-HumanID, and UT-HAR) demonstrate that Wi-Spike achieves competitive accuracy in single-action recognition and superior performance in multi-action recognition tasks. As for energy consumption, Wi-Spike reduces the energy cost by at least half compared with other methods, while still achieving 95.83\% recognition accuracy in human activity recognition.  More importantly, Wi-Spike establishes a new state-of-the-art in WiFi-based multi-action HAR, offering a promising solution for real-time, energy-efficient edge sensing applications.

\end{abstract}

% Note that keywords are not normally used for peerreview papers.
\begin{IEEEkeywords}
WiFi sensing, Human action recognition, Spiking Neural Network, Low-power sensing.
\end{IEEEkeywords}

\IEEEpeerreviewmaketitle

\section{Introduction}

% \hfill August 26, 2015

Human action recognition (HAR) has become a cornerstone of pervasive computing, enabling applications, such as smart healthcare, security monitoring, and human–computer interaction. Traditional HAR approaches rely on vision-based or wearable sensors, but these methods face inherent limitations: wearable devices require user compliance, while cameras raise privacy concerns and often depend on costly installation and line-of-sight availability. These challenges highlight the need for alternative solutions that are more practical, non-intrusive, and scalable.

WiFi-based action recognition problems \cite{liu2019wireless,ahmad2024wifi,tan2022commodity} have emerged as a promising direction. Usually, WiFi signals are inexpensive and ubiquitous, as wireless access points and routers are already widely deployed in homes and public spaces \cite{ge2022contactless}. This allows HAR systems to be built upon existing infrastructure without requiring specialized hardware. Moreover, WiFi signals penetrate walls and obstacles \cite{abuhoureyah2024wifi}, making it possible to monitor activities in non-line-of-sight conditions where vision-based systems fail. A key advantage of WiFi sensing lies in its non-intrusive and privacy-preserving nature \cite{bocus2021uwb, zhang2024wifi}. Unlike cameras, WiFi does not capture visual appearance, and unlike wearables, it does not require users to carry or maintain devices. These characteristics make WiFi particularly suitable for long-term daily monitoring \cite{li2025consense} in sensitive scenarios, such as hospitals, elderly care, or private households. Taken together, WiFi-based HAR combines ubiquity, low cost, privacy preservation, and ease of deployment, making it an important and valuable research direction. Its advantages over traditional methods position it as a key enabler for next-generation intelligent environments.

Early research on WiFi-based human action recognition (HAR) primarily relied on traditional signal processing and handcrafted feature extraction. Representative examples include WiSee, which first demonstrated the feasibility of leveraging Doppler shifts in commercial WiFi signals for through-wall gesture recognition \cite{adib2013see}; E-eyes, which constructed activity fingerprints from CSI/RSSI profiles to identify daily activities in device-free settings \cite{wang2014eyes}; and CARM, which built an analytical model mapping CSI variations to human motion speed profiles for activity recognition \cite{wang2015understanding}. These studies established the theoretical foundation of WiFi sensing but were limited by environmental sensitivity and the lack of scalable feature representation. In recent years, the field has been rapidly advanced by deep learning approaches, which automatically learn discriminative representations from CSI without requiring manual feature design. CNN-based models such as SignFi \cite{ma2018signfi} demonstrated effective recognition of sign language gestures by converting CSI into time–frequency images and applying convolutional feature extraction. RNN-based methods, such as the attention-enhanced BiLSTM framework \cite{chen2018wifi}, improved performance by modeling temporal dependencies while focusing on informative subcarriers. More recently, attention and Transformer-based architectures have emerged, enabling the capture of long-range dependencies in CSI sequences and achieving superior generalization across users and environments \cite{luo2024vision}. In addition, hybrid models that integrate CNN and attention mechanisms have further improved robustness and recognition accuracy in complex real-world settings \cite{shi2025cit}. These deep learning approaches have significantly boosted recognition accuracy and robustness, driving rapid progress in WiFi human action sensing research. The aforementioned studies on WiFi-based human action recognition primarily focus on simple and singular action, such as running, jumping, and walking. There is a lack of analysis on more complex and multi-action patterns.

Multi-action sensing problem refers to the ability to recognize multiple human actions or complex combinations of actions from WiFi signals. It extends the scope of existing research on simple action recognition, aiming to better capture and analyze more natural and complex human behaviors in daily life. Current WiFi-based sensing studies mainly focus on single actions, such as walking or jumping, or a limited set of predefined actions within a sequence. However, in real-world scenarios, human activities are often composed of multiple sub-actions with intricate temporal and spatial dependencies. Recognizing such composite actions poses significant challenges, including ambiguous transitions, inter-action interference, and multi-person overlap. Advancing research on multi-action sensing can enhance the capability of WiFi-based systems to understand complex behaviors, support a wider range of practical applications, and improve both adaptability and recognition accuracy. To address multi-action sensing problem, there are several promising solutions for recognizing complex multi-actions WiFi CSI signals. An early study \cite{zhang2022wi} explored the effective recognition of multiple gesture actions from CSI signals. Subsequent works further investigated the recognition and detection of simultaneous human actions \cite{liu2025wifi}, as well as the multi-action pattern recognition \cite{huang2024wimans} in multi-person scenarios. Overall, research on multi-action sensing with WiFi signals is still in its early stage.

\begin{figure*}
\centering % 表示居中
\includegraphics[width=0.98\textwidth]{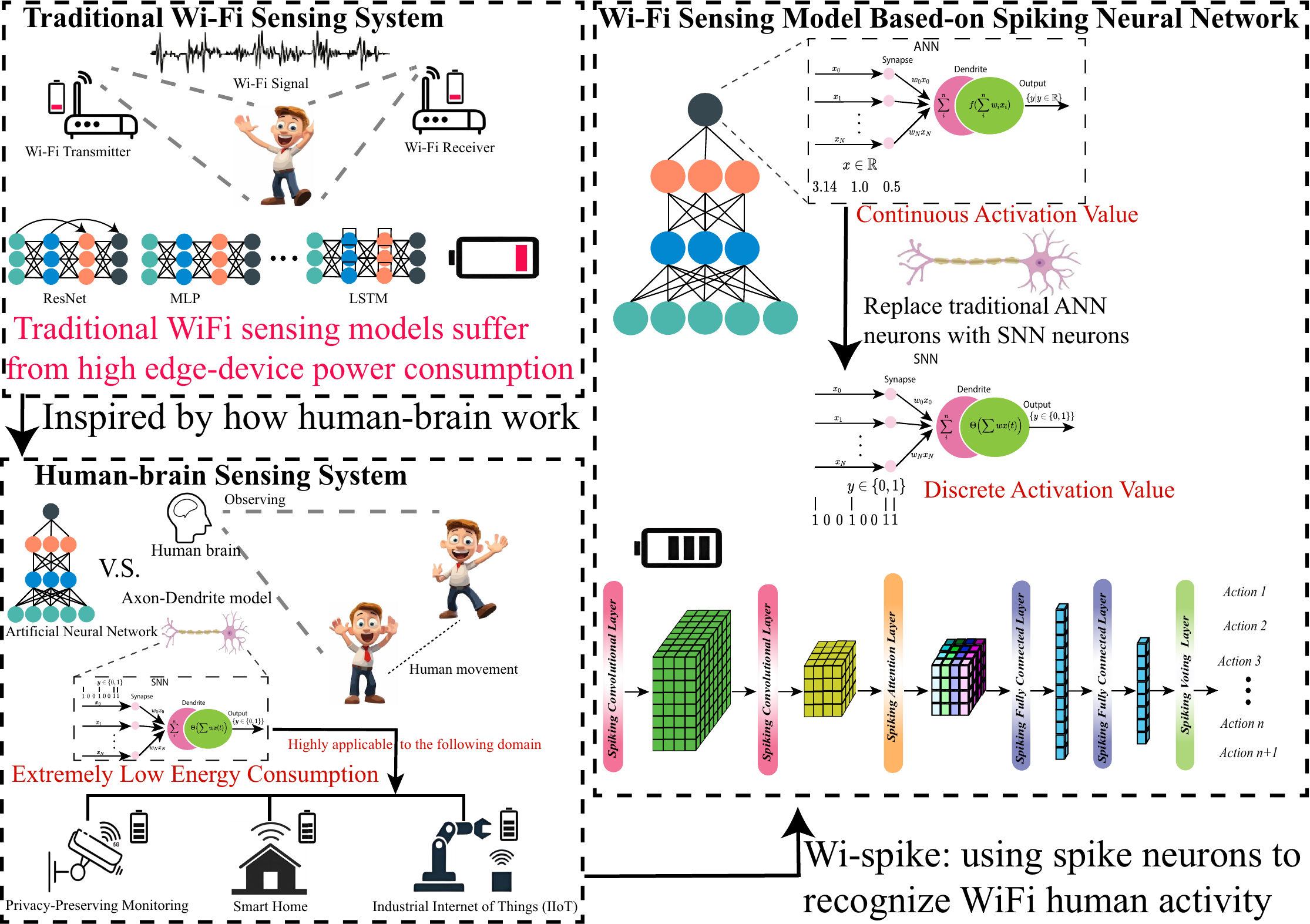}
\caption{ Overview of WiFi sensing model based on Spiking Neural Network. Conventional WiFi-based human sensing systems, relying on artificial neural networks, suffer from high energy consumption. Inspired by brain-like models, we propose a novel WiFi human action recognition model by replacing traditional neurons with spiking neurons. This approach,termed Wi-Spike, achieves a low-power WiFi-based human action recognition system.} \label{fig:WiFi_motivation}
\end{figure*}

Despite the rapid progress of WiFi-based sensing techniques, several important challenges remain unresolved. First, most existing studies primarily emphasize improving model accuracy and, to some extent, inference speed, yet they provide only limited consideration of inference power consumption. In scenarios requiring continuous monitoring, high power demand becomes a critical bottleneck that undermines the sustainability and practicality of WiFi sensing systems. Second, WiFi sensing models are often expected to operate on resource-constrained edge devices, such as IoT nodes or embedded platforms \cite{kong2022edge}, where computational resources and battery capacity are limited. Conventional deep learning models, with their heavy computational cost and high energy demand, are difficult to deploy in such environments, making them unsuitable for long-term activity monitoring. Third, there is still a lack of a unified WiFi sensing framework that can simultaneously guarantee recognition accuracy and ensure energy efficiency. Existing approaches tend to treat accuracy and efficiency as conflicting objectives, leading to designs that optimize one at the expense of the other. These limitations highlight a clear research gap: current WiFi sensing solutions cannot fully meet the dual requirements of high-performance recognition and energy-efficient operation in realistic edge scenarios. Addressing this gap calls for a fundamentally different design paradigm, one that integrates accuracy and energy efficiency within the same framework. Motivated by this challenge, this work introduces a bio-inspired spiking neural network model, Wi-Spike, which leverages the event-driven and low-power characteristics of SNNs to achieve efficient and accurate WiFi-based activity recognition (As shown in Fig. \ref{fig:WiFi_motivation}).

Spiking neural networks (SNNs) have recently emerged as a promising alternative to conventional deep learning models due to their temporal dynamics, and low-power operation. Unlike traditional architectures that rely on dense and continuous computations, SNNs process information in the form of discrete spikes, making them inherently energy-efficient and well-suited for real-time sensing on resource-constrained edge devices. This unique property makes SNNs a compelling choice for WiFi-based human action recognition, where continuous monitoring and low-energy operation are crucial. 

To the best of our knowledge, we are the first work to introduce a bio-inspired spiking neural network framework for WiFi action recognition, termed Wi-Spike. The proposed model leverages the event-driven characteristics of SNNs to efficiently extract and represent discriminative features from WiFi CSI signals. Specifically, Wi-Spike employs spiking convolutional layers to capture spatial and temporal patterns embedded in CSI data. To further enhance the model’s representational power, we design a novel temporal attention mechanism, which adaptively emphasizes salient activity-related features while suppressing irrelevant information. Different from conventional deep learning approaches, Wi-Spike exhibits a stronger ability to disentangle complex and dense WiFi motion patterns, owing to the temporal coding and sparse firing properties of SNNs. Finally, the extracted features are encoded through spiking fully connected layers and aggregated by a spiking voting layer for robust classification. This bio-inspired design not only ensures Wi-Spike as a novel and effective solution for energy-efficient WiFi sensing. Therefore, the contributions of this work are summarized as follows:
\begin{enumerate}
\item We propose a novel WiFi sensing model, termed Wi-Spike, for human action recognition. Unlike conventional WiFi sensing approaches based on deep neural networks, Wi-Spike leverages spike-based encoding to recognize CSI signals. Importantly, it achieves lower power consumption while maintaining competitive recognition accuracy in human action sensing. 
\item We introduce the multi-action WiFi sensing problem as a new challenge for human action recognition and demonstrate the effectiveness of Wi-Spike in addressing it. Compared with traditional WiFi-based recognition tasks, multi-action CSI signals exhibit more complex and diverse patterns. To evaluate this challenge, we design two-actions and three-actions datasets, and experimental results show that Wi-Spike consistently outperforms existing WiFi sensing models in capturing such composite action patterns.
\item To further enhance feature extraction, we design a spike-based temporal attention layer. This module enables Wi-Spike to capture salient spatio-temporal patterns of human activities from encoded CSI signals, thereby allowing spiking neural networks to learn more robust representations for action recognition.
\end{enumerate}

The remainder of this paper is organized as follows. Section \ref{2} presents related works on bio-inspired pattern recognition algorithms and WiFi signal recognition models in recent years. The basic concepts and fundamental theories of WiFi signals are described in Section \ref{3}. The proposed WiFi action recognition model (Wi-Spike) is presented in Section \ref{4}. Besides, The specific experimental results and the conclusion are described in Section \ref{5} and \ref{6}, respectively.

\section{Related work}\label{2}
This section describes several key related studies. Firstly, we present the development of bio-inspired pattern recognition algorithms. Then, we lists recent some works related to human action recognition using WiFi signal. Next, the multi-action WiFi sensing techniques are described.

\subsection{Spiking Neural Network for WiFi Pattern Recognition}
Bio-inspired pattern recognition has gained significant attention in recent years, with Spiking Neural Networks (SNNs) \cite{ghosh2009spiking} emerging as a promising alternative to traditional deep learning models. Unlike conventional artificial neural networks (ANNs), SNNs mimic the event-driven processing of biological neurons, encoding information through discrete spikes rather than continuous activations. A growing body of research has demonstrated the potential of SNNs in pattern recognition tasks. A work \cite{luteijnspiking} showed the feasibility of applying SNNs to gesture recognition using WiFi CSI sensing, highlighting their ability to exploit temporal signal dynamics. Another work \cite{wang2024self} introduced a self-attention spiking neural network for human interaction recognition, which demonstrated the effectiveness of integrating attention mechanisms into spike-driven models. Subsequently, a study \cite{lee2023wi} further advanced the field by leveraging memristive synapses in SNNs for WiFi frame detection, showing the promise of hardware-efficient implementations.
An energy-efficient wireless technology recognition method was proposed in \cite{hu2025energy} by combining time-frequency feature fusion with spiking networks, and it was proven that complex wireless signals can be effectively processed by SNNs. In addition, recent works \cite{fang2021deep,shi2024spikingresformer} in vision-based applications—such as spiking convolutional architectures for image classification—have shown that hybrid designs combining convolution and spike-based computation can capture spatiotemporal dependencies while maintaining low-power consumption.

\begin{figure}
		\centering % 表示居中
		\includegraphics[width=0.5\textwidth]{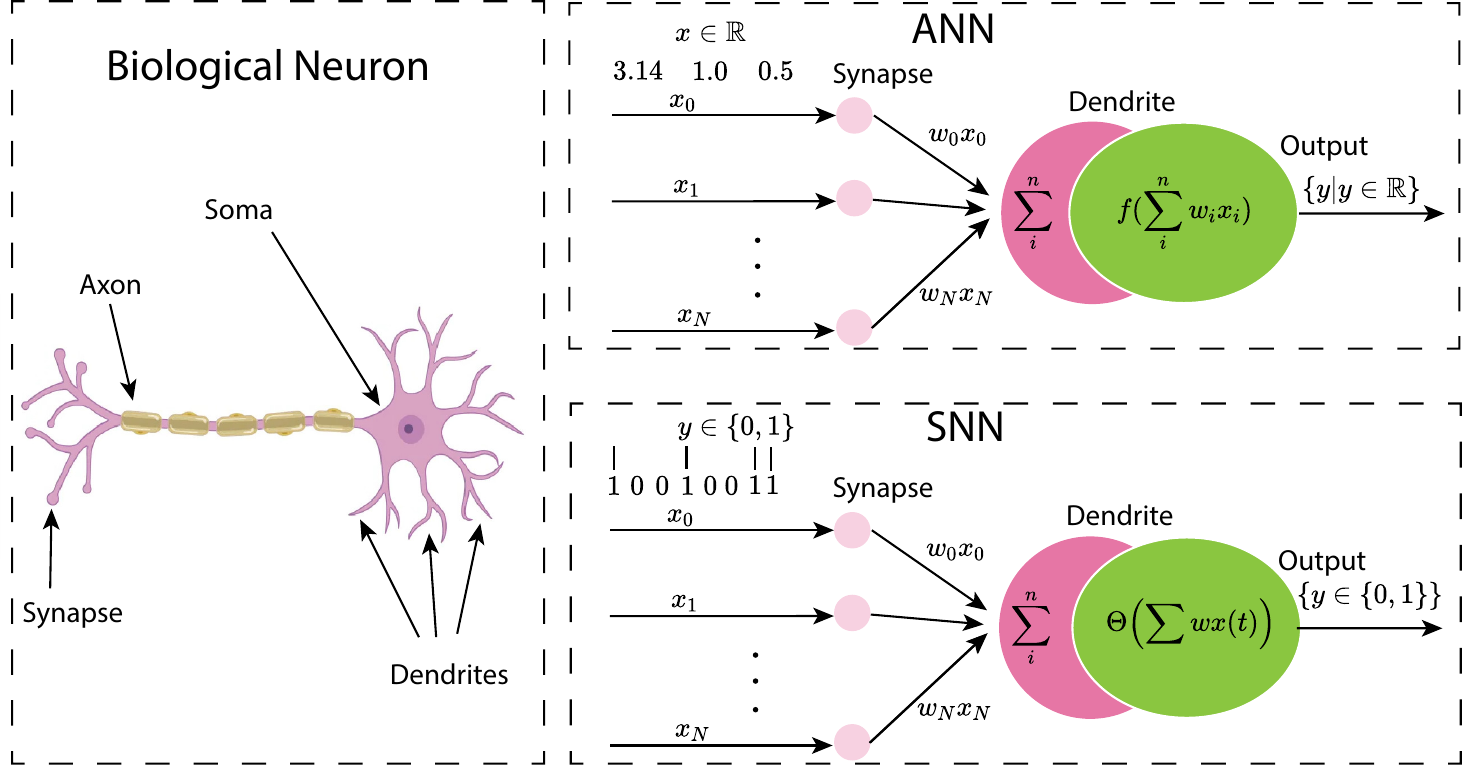}
		\caption{Description of the basic principles of Spiking Neural Networks and Artificial Neural Network.} \label{fig:2D_motion}
\end{figure}

\begin{figure*}[t] 
		\centering % 表示居中
		\includegraphics[width=1\textwidth]{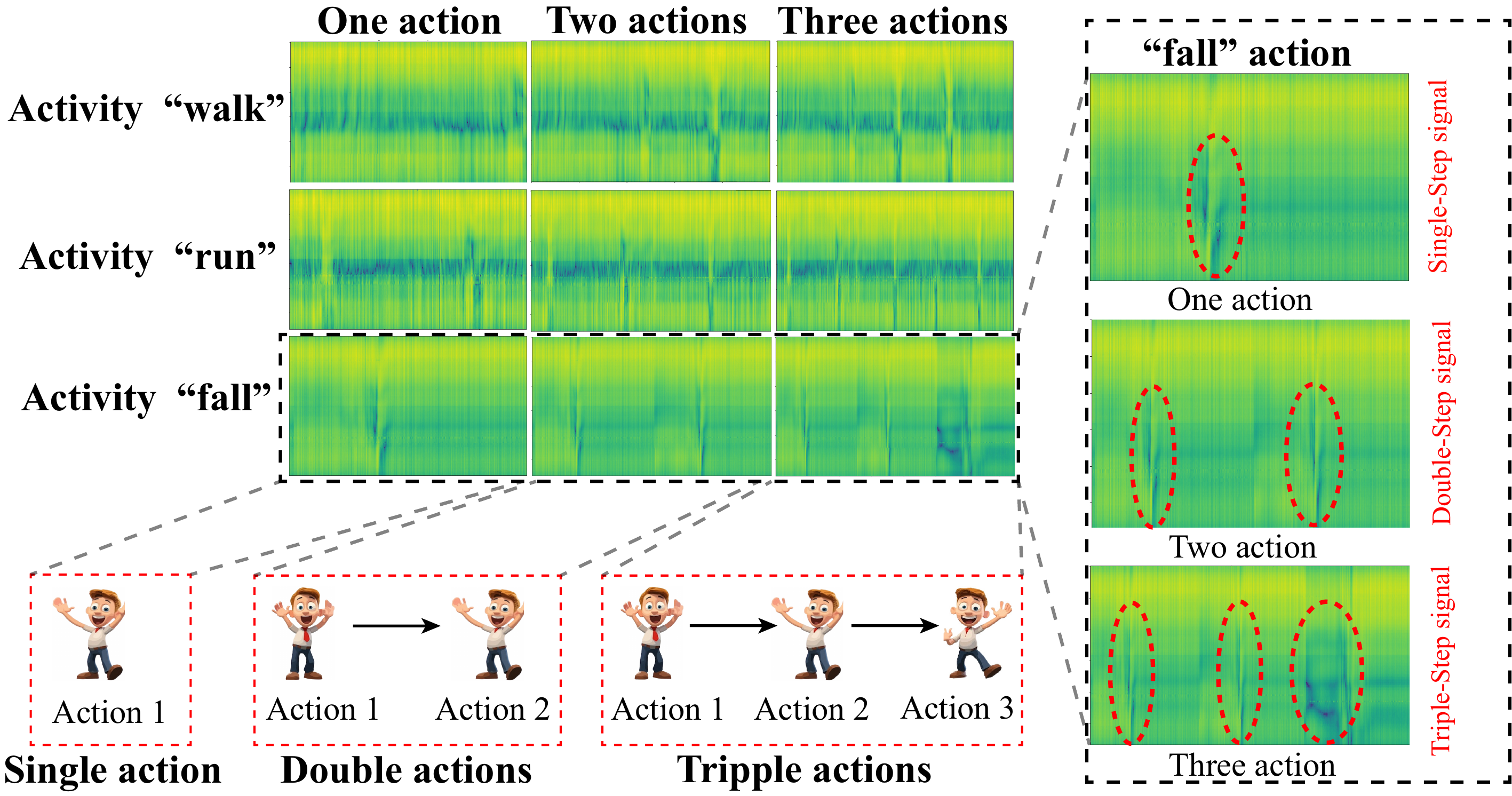}
            
		\caption{ A detailed comparison is provided in the single-action WiFi recognition and multi-action WiFi recognition problems. This includes WiFi CSI signals for one action, two actions, and three actions. In the right part of Fig. \ref{fig:dense}, clearly, the more human motion patterns in the WiFi CSI, the greater the number of step signals.} 
        \label{fig:dense}
\end{figure*}

\subsection{The Development of WiFi sensing techniques}
Human action recognition using WiFi signals has attracted increasing attention due to its ability to enable device-free sensing, privacy-preserving monitoring, and robust operation under poor lighting or occluded conditions. Unlike vision-based approaches, WiFi-based sensing leverages channel state information (CSI) or raw signals to capture body movements, offering a promising solution for smart homes, healthcare, and human-computer interaction. Early works mostly relied on handcrafted features and statistical modeling, such as Doppler shift, CSI variance, or frequency analysis. While these methods provided initial insights, they often suffered from limited generalization ability and poor robustness to environmental dynamics. Consequently, recent research has shifted toward deep learning-based approaches, which automatically extract discriminative spatiotemporal features and significantly outperform traditional counterparts.

\begin{figure*}[t]
		\centering % 表示居中
		\includegraphics[width=0.98\textwidth]{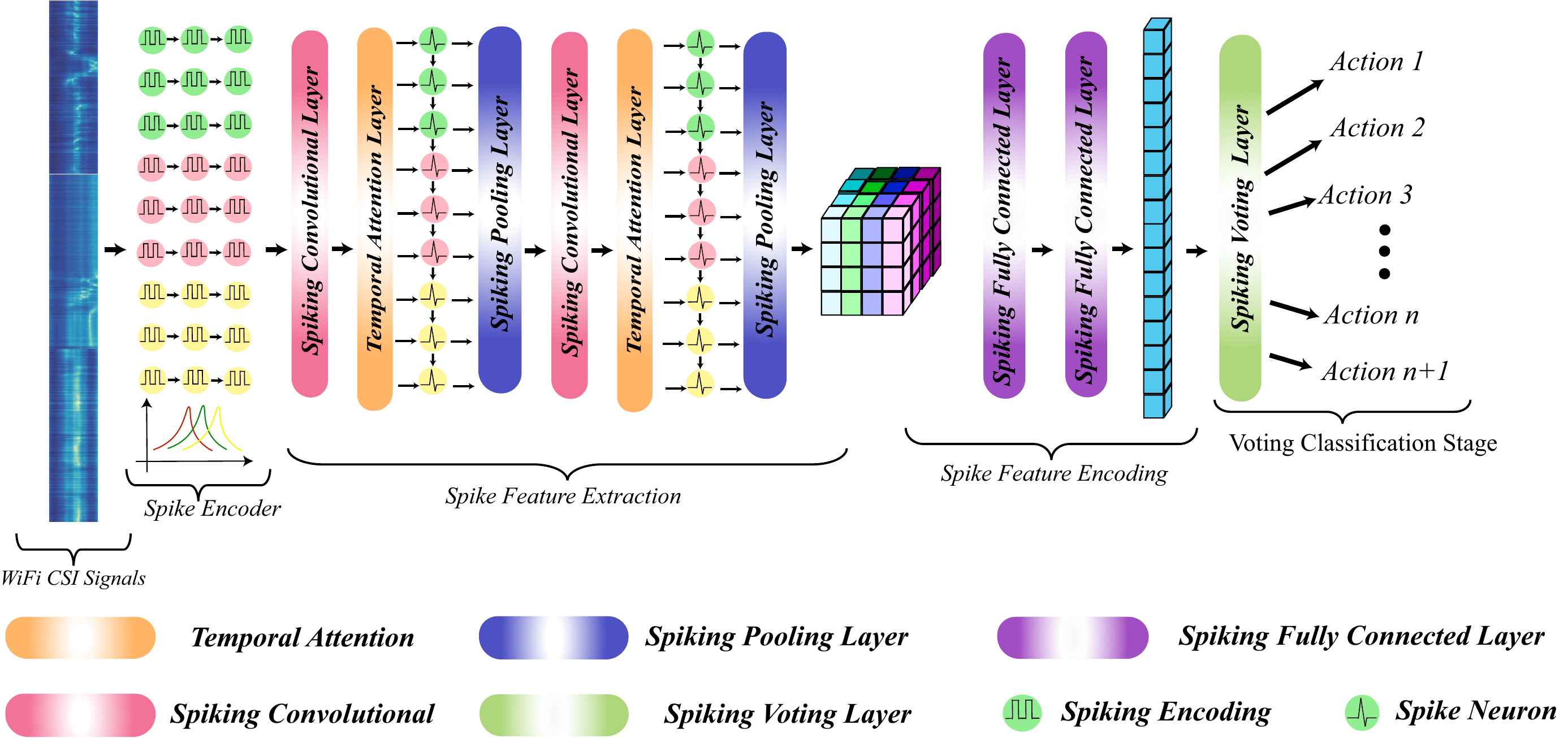}
            
		\caption{ Overview of Wi-Spike Architecture. From left to right, the raw WiFi CSI human activity signals sequentially pass through the spike encoder, spike feature extraction, spike feature encoding, and voting classification modules.} \label{sample-figure}
\end{figure*}

There are some works that learn spatially discriminative representations from CSI data. CSI-Net \cite{wang2018csi} pioneered the use of end-to-end CNNs by transforming raw CSI measurements into image-like tensors and applying convolutional feature extraction. Similarly, SignFi \cite{ma2018signfi} converted CSI streams into time–frequency spectrograms and achieved effective recognition of sign language gestures, demonstrating the utility of convolutional architectures in capturing localized spatial–frequency features. More recently, a work \cite{el2024csi} using a lightweight CNN to achieve real-time action recognition without sacrificing accuracy. Besides, Recurrent neural networks (RNNs) have been extensively studied to capture temporal dependencies in CSI sequences. A representative example is the attention-enhanced BiLSTM framework \cite{chen2018wifi}, which improved recognition performance by focusing on informative subcarriers while modeling bidirectional temporal context. Widar \cite{zhang2021widar3} further leveraged recurrent architectures for robust gesture recognition, demonstrating that RNNs excel in handling fine-grained motion dynamics. More recent studies have explored attention and Transformer-based models, which are particularly suited for capturing long-range dependencies and contextual information in CSI sequences. A study \cite{luo2024vision} introduced the transformer model to identify WiFi CSI signals and analyzed the computational efficiency and model size of the attention model. CIT-HAR \cite{shi2025cit} fuses CNN feature extractors with Transformer layers to achieve both robustness and model compactness.

\subsection{Multi-action WiFi Sensing Techniques }

The multi-action WiFi sensing technology has developed rapidly in recent years, utilizing CSI signals to capture subtle disturbances from human movements for non-contact recognition. This field has evolved from early focus on single-user multi-gesture sequences to continuous action detection and multi-user scenarios, forming a clear technological progression: The earliest Wi-Phrase work \cite{zhang2022wi} introduced a deep residual network combined with multi-head attention mechanisms for WiFi-based sign language phrase recognition, extracting robust features via residual modules and capturing spatio-temporal relationships between gestures to successfully sense multiple continuous gestures, laying the foundation for transitioning from isolated actions to sequence recognition; Subsequently, DPWiT \cite{liu2025wifi} proposed a dual pyramid network architecture that separates high- and low-frequency CSI features and fuses them through cross-attention for end-to-end temporal activity detection, locating and classifying multiple continuous actions in untrimmed signals, advancing the technology toward long-duration multi-actions; In recent years, WiMANS \cite{huang2024wimans} created the first multi-user WiFi benchmark dataset, covering dual-band CSI samples and synchronized videos to support simultaneous action recognition for 0-5 users, benchmarking existing models and emphasizing multi-modal fusion, achieving a leap from single-user multi-actions to multi-user parallel behaviors. This progression shows the technology is evolving toward more complex real-world scenarios. However, current methods heavily rely on traditional deep learning, facing challenges in power consumption and real-time performance. There is still a lack of research on low-power WiFi sensing technology.

\section{Preliminary study and problem statements} \label{3}
This section presents core foundational concepts, principally comprising an introductory exposition of WiFi Channel State Information (CSI) and the formal definition of multi-action recognition using WiFi signals. Lastly, the principles of Spiking Neural Network are introduced. 

\subsection{WiFi Channel State Information (CSI) Signal}
WiFi Channel State Information (CSI) provides a fine-grained representation of wireless channel characteristics by capturing the amplitude and phase of subcarriers in Orthogonal Frequency Division Multiplexing (OFDM)-based systems. Due to its sensitivity to environmental changes caused by human presence or motion, CSI has been widely applied in indoor localization, action recognition, and gesture detection. In off-the-shelf WiFi systems, CSI describes the Channel Frequency Response (CFR) of each subcarrier. For a Multiple-Input Multiple-Output (MIMO) system with $N_t$ transmit and $N_r$ receive antennas, the received signal at the $k$-th subcarrier can be expressed as:
\begin{equation}
\mathbf{Y}_k = \mathbf{H}_k \mathbf{X}_k + \mathbf{N}_k,
\label{eq:csi_model}
\end{equation}
where $\mathbf{Y}_k \in \mathbb{C}^{N_r \times 1}$ is the received signal, $\mathbf{X}_k \in \mathbb{C}^{N_t \times 1}$ is the transmitted signal, $\mathbf{H}_k \in \mathbb{C}^{N_r \times N_t}$ is the CSI matrix, and $\mathbf{N}_k \in \mathbb{C}^{N_r \times 1}$ denotes additive white Gaussian noise.

For a transmit-receive antenna pair $(i,j)$, the CSI at subcarrier $k$ is:
\begin{equation}
H_k^{i,j} = |H_k^{i,j}| e^{j \angle H_k^{i,j}},
\label{eq:csi_amplitude_phase}
\end{equation}
where $|H_k^{i,j}|$ represents amplitude attenuation due to path loss and multi-path fading, and $\angle H_k^{i,j}$ denotes phase shifts introduced by propagation. These properties make CSI a powerful indicator of environmental dynamics and the foundation of WiFi-based sensing.

\subsection{The definition of WiFi-Based Human Multi-action Recognition}

In wireless sensing, WiFi-based action recognition exploits Channel State Information (CSI) to perceive human activities by capturing signal variations induced by body movements. When a person performs an action, such as walking, waving, or turning, the resulting changes in multi-path propagation and signal reflection patterns are embedded in the CSI measurements. By analyzing these fine-grained perturbations, it becomes possible to detect and classify human actions without requiring cameras or wearable sensors, enabling non-intrusive and device-free action recognition. Besides, to more clearly demonstrate multi-action WiFi-based human action recognition, Fig. \ref{fig:dense} provides a detailed visualization of WiFi CSI signals for varying numbers of human actions.

When a single human action $A$ is performed in front of a WiFi device, 
the observed complex Channel State Information (CSI) over the observation window $[0,T]$ can be written as in Eq. \ref{eq:csi_single_action_mod}.
\begin{equation}
\begin{split}
& \mathbf{H}_A(t) = \big\{ H_k^{i,j}(t;A) \ \big|\ i=1,\dots,N_t,\ \\
& j=1,\dots,N_r,\ k=1,\dots,K \big\}, \quad t \in [0,T],
\label{eq:csi_single_action_mod}
\end{split}
\end{equation}
where $H_k^{i,j}(t;A)\in\mathbb{C}$ is the CSI coefficient of the $k$-th OFDM subcarrier between transmit antenna $i$ and receive antenna $j$ at time $t$, explicitly parameterized by the action $A$. Here $N_t$, $N_r$ and $K$ denote the numbers of transmit antennas, receive antennas and subcarriers, respectively. A single action instance is thus defined as the action-specific perturbation contained in $\mathbf{H}_A(t)$. For compactness one can introduce a simple front-end operator $\phi(\cdot)$ that extracts an action-induced perturbation signal in Eq. \ref{eq:delta_A}.
\begin{equation}
\delta_A(t)=\phi\big(\mathbf{H}_A(t)\big),\qquad t\in[0,T],
\label{eq:delta_A}
\end{equation}
where $\delta_A(t)$ (e.g., magnitude/phase difference or short-time energy) summarizes the amplitude/phase variations of the CSI caused by multi-path modulation and Doppler effects induced by the action implementing $A$.

While single-action WiFi recognition assumes independent activities within clear temporal windows, real-world scenarios often involve continuous or overlapping movements. Both rely on analyzing perturbations in $\mathbf{H}(t)$, but multiply action recognition differs in that $\mathbf{H}(t)$ is influenced by multiple successive actions, producing entangled patterns without explicit boundaries and thus making recognition more challenging.

Let $\mathcal{A}=\{A_1,\dots,A_M\}$ denote the (latent) sequence of $M$ atomic actions within $[0,T]$, and let $y^\star\in\mathcal{C}_{\mathrm{comp}}$ denote a \emph{composite class} formed by a particular combination of these atomic actions. The observed CSI stream is $\mathbf{H}(t),\ t\in[0,T]$. Using the same front-end operator $\phi(\cdot)$ as in the single-action definition, define the action-induced perturbation signals in Eq. \ref{eq:delta_defs}.
\begin{equation}
\begin{split}
& \delta_A(t)=\phi\big(\mathbf{H}_A(t)\big),\qquad \\
& \delta_{y^\star}(t)=\phi\big(\mathbf{H}(t)\big),\quad t\in[0,T],
\label{eq:delta_defs}
\end{split}
\end{equation}
where $\delta_A(t)$ is the perturbation produced by a single atomic action $A$, and $\delta_{y^\star}(t)$ is the perturbation observed when multiple atomic actions combine into the composite class $y^\star$.

The multi-action (composite-class) recognition task is to map the CSI stream to a composite label:
\begin{equation}
\begin{split}
& f_{\mathrm{comp}}:\ \mathbf{H}(t) \mapsto y^\star\in\mathcal{C}_{\mathrm{comp}},\quad \\
& y^\star=\arg\max_{c\in\mathcal{C}_{\mathrm{comp}}}P\big(c\mid\mathbf{H}(t)\big),\\
& N_x \triangleq \int_{0}^{T}\mathbf{1}\{\delta_x(t)>\varepsilon\}\,dt,\qquad \\
& N_{y^\star}\gg N_{A},
\end{split}
\label{eq:comp_map_density}
\end{equation}
where $f_{\text{comp}}$ denotes the classifier for multi-action recognition, which maps $H(t)$ to a target class $y^* \in C_{\text{comp}}$ by maximizing the posterior probability $P(c \mid H(t))$. Besides, $\varepsilon$ is a small threshold and $N_x$ measures the effective activity density of $\delta_x(t)$. The relation $N_{y^\star}\gg N_A$ formalizes the key distinction: a composite-class signal $\delta_{y^\star}(t)$ is substantially more \emph{dense} (contains more effective pulses) than the perturbation $\delta_A(t)$ produced by a single human action, which is the source of the increased recognition difficulty in multi-action scenarios.

\subsection{The Basic Principles of Spiking Neural Network}
The basic principle of an SNN neuron involves integrating incoming signals until a threshold is reached, at which point the neuron emits a spike. Specifically, the vanilla version of Spiking Neural Networks (SNNs) is composed of a series of neurons in Eq. \ref{eq:snn}. In Fig. \ref{fig:2D_motion}, we present the schematic diagram of the principle and structure of spiking neural networks (SNNs), along with a detailed comparison of their computational logic with Artificial Neural Networks (ANNs).
Taking the Leaky Integrate-and-Fire (LIF) neuron \cite{stoliar2017leaky} as an example, spikes are defined as binary states $S(t) \in {0,1}$. A complete description of the computational process of the LIF neuron is given in Eqs. \ref{eq:snn} and \ref{ed:snnLIF}. 

\begin{equation}
\begin{split}
  & V(t+1) = V(t) \cdot e^{-\frac{\Delta t}{\tau_m}} + R \cdot I(t), \\
 &  I(t) = \sum_j{w_j \cdot S_j(t)}. 
\end{split}
\label{eq:snn}
\end{equation}
The membrane potential $V(t)$ evolves over time, influenced by input currents and leak dynamics, where $\tau_m$ is membrane time constant and $R$ represents membrane resistance. Lastly, $I(t)$ is external input current, which is also named as weighted sum of incoming spikes. 

\begin{equation}
\begin{split}
& S(t) =
\begin{cases} 
1, & \text{if } V(t) \geq V_{th}, \\
0, & \text{otherwise}.
\end{cases}
\end{split}
\label{ed:snnLIF}
\end{equation}
In Eq \ref{ed:snnLIF}, $S(t)$ is the spike generation process. When $V(t)$ exceeds a certain threshold, a spike is triggered.   

\section{The Proposed Method}\label{4}
This section firstly presents the overview of our proposed model. Then, we show CSI pre-processing and encoding techniques. Next, spiking temporal attention and spiking voting layer are described. Lastly, the final training objectives are given in this part.

\subsection{The overview of Wi-Spike Model}
The proposed Wi-Spike model leverages WiFi Channel State Information (CSI) signals as the input, 
initiating the process with a Spike Encoder that transforms continuous CSI data into discrete spike trains. 
This step emulates the event-driven nature of biological neurons, enabling efficient processing of 
temporal dynamics inherent in human activities. The encoded spikes then flow into the Spike Feature 
Extraction stage, where spiking convolutional layers and a novel temporal attention mechanism extract 
and enhance spatial and temporal features. The attention mechanism, inspired by human cognitive focus, 
prioritizes discriminative patterns, significantly improving the model's ability to distinguish complex multi-action scenarios.

The extracted features are subsequently passed to the Spike Feature Encoding stage, where spiking pooling layers reduce spatial dimensions while preserving critical temporal information. This process aligns with the low-power, event-driven advantages of spiking neural networks (SNNs). The encoded representations are then fed into the Voting Classification Stage, comprising spiking fully connected layers and a spiking voting layer, which collectively classify the activities by aggregating spike-based decisions, achieving robust performance in both WiFi single- and multi-action recognition tasks.

\subsection{WiFi CSI Signal Pre-processing and Spike Encoding}
Our Wi-Spike model processes raw WiFi CSI signals through pre-processing, spike encoding to capture spatial-temporal patterns for human action recognition.
Given a raw WiFi CSI matrix $x \in \mathbb{R}^{C \times H \times W}$, we first apply mean subtraction for normalization to remove global bias \cite{aravindan2024comparative}:
\begin{equation}
\begin{split}
\tilde{x} = x - \mu, \quad \mu = \frac{1}{C \cdot H \cdot W} \sum x.
\end{split}
\label{ed:precess}
\end{equation}
Next, for spike encoding, the preprocessed $\tilde{x}$ is expanded to the time dimension by duplicating it over $T$ time steps, creating a temporal input tensor $X^T \in \mathbb{R}^{T \times C \times H \times W}$:
\begin{equation}
\begin{split}
X^T = \tilde{x} \otimes \mathbf{1}_T,
\end{split}
\label{ed:expand}
\end{equation}
where $\mathbf{1}_T$ is a vector of ones with length $T$, effectively replicating $\tilde{x}$ along the new time axis. This constant-rate encoding treats the static CSI features as sustained input currents over time, enabling temporal processing in the subsequent network layers. These CSI features are then extracted using a Convolutional Spiking Layer.

\subsection{Temporal Attention and Spiking Voting Layer}

\begin{figure}[t]
		\centering % 表示居中
		\includegraphics[width=0.49\textwidth]{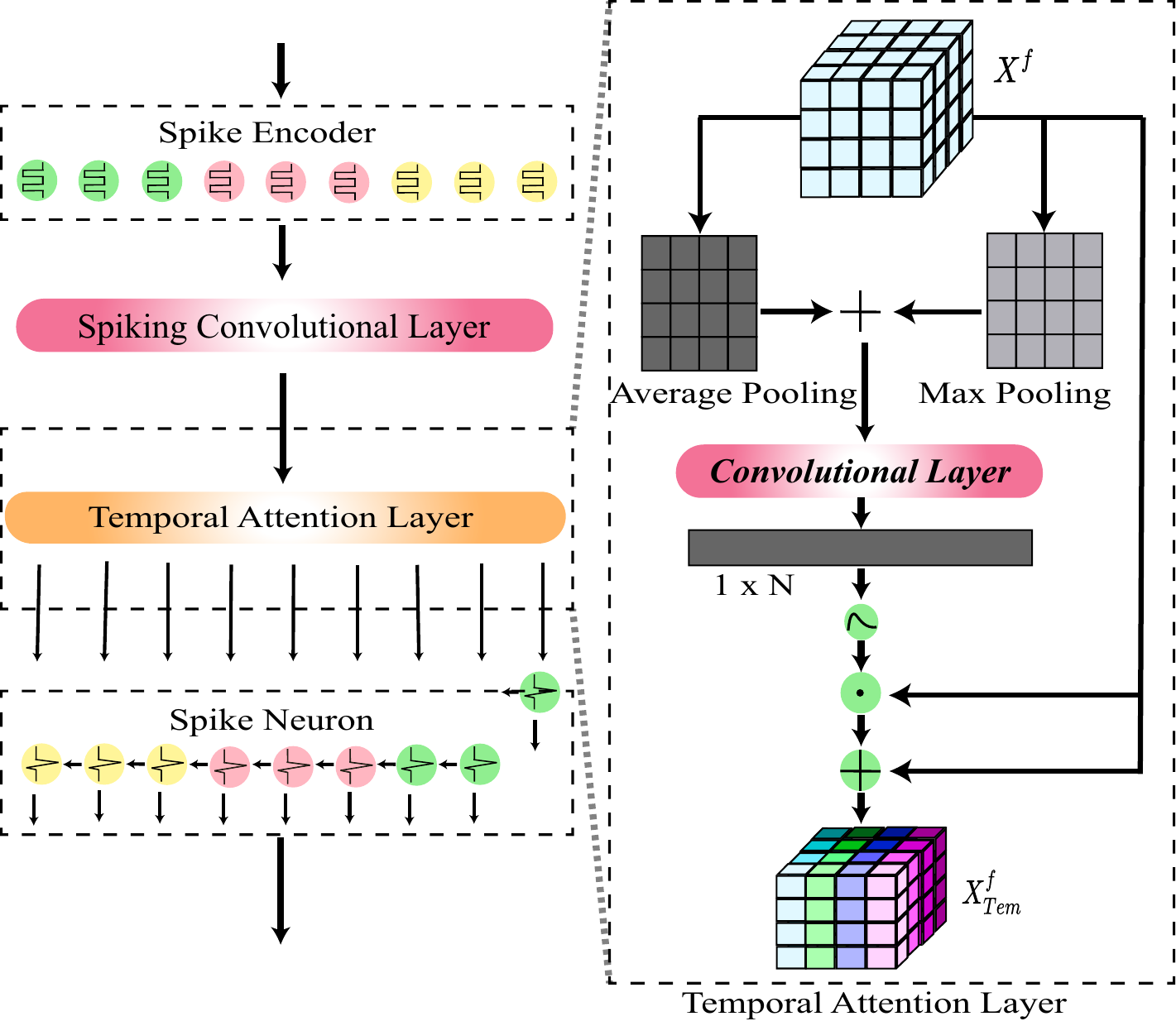}        
		\caption{Detailed structure of the Temporal Attention Layer. } \label{fig:FPN}
\end{figure}

In order to enhance human multi-action recognition, the Temporal Attention Layer enhances processing of time-varying CSI signals from WiFi transmissions. SNNs encode data via discrete spikes, offering energy efficiency for sparse, event-driven WiFi CSI changes due to movements like walking or falling. Integrated after spiking convolutional layers, this layer recalibrates temporal features to focus on key segments with discriminative Doppler shifts or amplitude variations in CSI. It benefits WiFi recognition by handling temporal dependencies and multipath noise, reducing irrelevant signals, amplifying behavior patterns, and maintaining SNN efficiency.
Mathematically, the layer first processes the feature map $X^f$ from prior layers using localized pooling and affine transformation:
\begin{equation}
\begin{split}
& f_{\text{avg}}(X^f)(i,j) = \frac{1}{k_h \cdot k_w} 
\sum_{m=0}^{k_h-1} \sum_{n=0}^{k_w-1} 
X^f(i \cdot s + m, j \cdot s + n), \\
& f_{\text{max}}(X^f)(i,j) = \max_{0 \leq m < k_h,\; 0 \leq n < k_w} 
X^f(i \cdot s + m, j \cdot s + n), \\
& M(i,j) = \alpha \cdot f_{\text{avg}}(X^f)(i,j) + \beta \cdot f_{\text{max}}(X^f)(i,j).
\end{split}
\end{equation}

Here, $X^f$ is the input tensor encoding spatio-temporal CSI features; $i, j$ are spatial indices; $k_h, k_w$ are pooling kernel sizes; $s$ is stride; $\alpha, \beta$ are learnable scalars balancing pooling; and $M$ is the descriptor capturing combined pooled information. Subsequently, the layer computes global recalibration to derive attention weights and the final output:
\begin{equation}
\begin{split}
& M_{\text{mean}} = mean(M), \\
& T_{w} = Sigmoid(Linear(M_{\text{mean}})),  \\
& X^f_{Tem} = X^f + X^f \cdot T_{w}. 
\end{split}
\end{equation}

Here, $M_{\text{mean}}$ is the global mean of $M$; $\text{Linear}(\cdot)$ is a fully connected layer for transformation; $\text{Sigmoid}(\cdot)$ generates weights in [0,1]; $T_w$ is the attention vector; and $X^f_{Tem}$ is the output for spiking neurons.
This layer adds residual attention for SNNs, weighting temporal CSI dimensions to highlight motion frames and suppress noise, improving discriminability, reducing overfitting, and enhancing accuracy in WiFi action recognition.

In the final classification stage of our Wi-Spike model, the Spiking Voting Layer serves as the final decoder, aggregating spike activities from preceding IF neurons to produce classification scores for human activities. It computes the population firing rate by averaging spikes over time, enabling a voting-like mechanism where the most active neuron group (corresponding to a class) determines the predicted action. Mathematically, the layer calculates the output as follows:

\begin{equation}
\begin{aligned}
f_c = \frac{1}{T} \sum_{t=1}^{T} z_t.
\end{aligned}
\end{equation}

Here, $f_c \in \mathbb{R}^n$ is the final output vector representing class scores, with $n$ denoting the number of activity classes; $T$ is the total number of time steps in the simulation; and $z_t \in \{0,1\}^n$ is the binary spike vector at time step $t$, where each element indicates whether the corresponding output neuron fired a spike.
In SNN-based WiFi action recognition, this layer integrates temporal spike patterns from CSI signals, providing a robust, noise-tolerant classification by emphasizing sustained neural activity over transient noise, thus enhancing ability in detecting dynamic human motions while leveraging the low-power, event-driven nature of spiking computations.

\subsection{The Training Objectives in the proposed Wi-Spike}
To obtain the optimal weights in our model, we employ a hybrid training loss that combines Mean Squared Error (MSE) with Supervised Contrastive Loss (SCL) to enhance both classification accuracy and feature discriminability. This hybrid approach is particularly effective in Wi-Spike, as it leverages the strengths of rate-based supervision from MSE while incorporating contrastive learning to promote better separation of activity representations in the spiking domain, ultimately leading to improved generalization and robustness in WiFi CSI-based human action recognition. The total loss function is defined as:
\begin{equation}
\begin{aligned}
\mathcal{L} = \gamma_1 \mathcal{L}_{\text{MSE}} + \gamma_2 \mathcal{L}_{\text{SCL}},
\end{aligned}
\end{equation}
where $\mathcal{L}_{\text{MSE}} = \frac{1}{N} \sum_{i=1}^{N} \left\| f_i - y_i \right\|^2$ is the MSE loss between the predicted firing rate $f_i$ (averaged over the time steps in the spiking output) and the one-hot ground truth label $y_i$ for each sample $i$ in a batch of size $N$; $\mathcal{L}_{\text{SCL}} = -\frac{1}{N} \sum_{i=1}^{N} \frac{1}{|P(i)|} \sum_{p \in P(i)} \log \left( \frac{\exp(\text{sim}(z_i, z_p) / \tau)}{\sum_{a=1, a \neq i}^{2N-1} \exp(\text{sim}(z_i, z_a) / \tau)} \right)$ is the supervised contrastive loss, computed on projected embeddings $z$ derived from the spiking features (using a small projection head), with $\text{sim}(\cdot, \cdot)$ denoting cosine similarity, $\tau$ as the temperature parameter (typically set to 0.07), $P(i)$ as the set of positive samples (i.e., other samples in the augmented batch sharing the same class label as $i$), and the denominator summing over all other samples in the batch (including augmented views); and $\gamma_1, \gamma_2 > 0$ are hyper-parameters that balance the contributions of the two loss terms (e.g., tuned via cross-validation to prioritize classification or representation learning as needed).

The MSE loss encourages the spiking output rates to align closely with target class distributions, facilitating reliable rate-based decoding in SNNs, while the SCL component pulls embeddings of similar activities closer and pushes dissimilar ones apart, enhancing the model's ability to capture fine-grained distinctions in WiFi CSI signals. This training phase is crucial for achieving state-of-the-art performance in multi-action recognition tasks, as it not only optimizes for accuracy but also ensures energy-efficient convergence by exploiting the sparse, event-driven nature of SNNs, making Wi-Spike suitable for deployment on edge devices.

\section{Experiments}\label{5}
In this section, we describe the key aspects of the experiment. First, we introduce the primary datasets used in this study. Next, we detail the experimental setup and parameter configurations. Then, we conduct human action recognition experiments for both single-action and multi-action WiFi signals in Section C and Section D. Then, the firing rate analysis and energy consumption comparison of Wi-Spike are presented in Section E. Finally, we visualize the feature maps of Wi-Spike and traditional CNN model in Section F.

\subsection{The Introduction of Experimental Datasets}
%UT-HAR \cite{yousefi2017survey}.

%  new generate segmentation
Three publicly available WiFi sensing datasets are used in our experiments: UT-HAR \cite{yousefi2017survey}, NTU-Fi-HumanID  \cite{yang2022efficientfi}, and NTU-Fi-HAR \cite{wang2022caution}. The UT-HAR dataset, collected using an Intel 5300 NIC with three antenna pairs, records 30 subcarriers per pair and includes seven activity categories. Data were gathered continuously in a single environment without golden labels for activity segmentation, and a sliding window was applied, resulting in approximately 5,000 samples with some repetition. Despite its size, UT-HAR has intrinsic limitations due to its segmentation approach. The NTU-Fi-HAR dataset, utilizing the Atheros CSI Tool with 114 subcarriers per antenna pair, covers six activities (box, circle, clean, fall, run, and walk) across three layouts, with 956 training and 264 testing samples, each sized (3,114,500) (antenna, subcarrier, packet). The NTU-Fi-HumanID dataset targets gait-based identification of 14 subjects under three scenarios (wearing a t-shirt, coat, or backpack), with 546 training and 294 testing samples of the same size. Both NTU-Fi datasets, segmented by threshold and time duration for complete data capture and trained over 30 epochs, along with UT-HAR, provide a robust and diverse foundation for evaluating deep-learning models on CSI data.

To evaluate the performance of models in recognizing human actions from WiFi CSI signals in multi-action scenarios, we constructed new datasets by recombining the UT-HAR, NTU-Fi-HumanID, and NTU-Fi-HAR datasets. Specifically, we concatenated two individual WiFi CSI action samples to form a new CSI signal, effectively increasing the number of actions within a single CSI sample. Following this approach, we generated the UT-HAR\_Double, NTU-Fi-HumanID\_Double and NTU-Fi-HAR\_Double datasets, each containing two human actions, as well as the UT-HAR\_Triple, NTU-Fi-HumanID\_Triple and NTU-Fi-HAR\_Triple datasets, each containing three human actions.

\subsection{Experimental Setup and Evaluation Metrics}
All experiments are conducted on a workstation equipped with an NVIDIA RTX 4090 GPU (48 GB memory) and 96 GB RAM running the Ubuntu operating system. The deep learning framework adopted was PyTorch version 2.7.0.
During training, the models were optimized using the Adam optimizer for approximately 30 epochs, with a batch size of 32 and an initial learning rate of 0.01. The loss function combined mean squared error (MSE) with a supervised contrastive learning strategy, which is well suited for the optimization characteristics of spiking neural networks (SNNs) based on membrane potential dynamics. For data preprocessing, WiFi CSI signals were directly loaded from .mat files and normalized. Each CSI sample was reshaped into a tensor of size 3×114×500, converted into a PyTorch FloatTensor, and optionally passed through simple transformations if specified. Unlike traditional approaches that rely on handcrafted preprocessing (e.g., Short-Time Fourier Transform, STFT), the proposed framework leverages the raw CSI matrices directly as network inputs, thereby avoiding manual feature engineering and preserving the intrinsic temporal-spatial patterns of the signal. Each input sample was treated as an entire instance, and the model was trained to predict one action label per sample, corresponding to holistic classification rather than frame-wise predictions. This setting enables the model to exploit the global temporal structure of the CSI data for robust action recognition.

In addition, we compare our approach with a series of representative WiFi-based human action recognition models. Specifically, the benchmarks include an LSTM model that extracts temporal features from WiFi CSI signals, a ViT model capable of capturing long-range contextual dependencies, and CNN- and ResNet18-based models that learn spatial-domain representations of CSI signals. Moreover, we further evaluate a hybrid CNN+GRU model that integrates both temporal and spatial information for action recognition.

Besides, in the context of SNNs, the firing rate serves as a critical metric. It quantifies the frequency at which neurons emit spikes (discrete action potentials) within a given time window. This measure reflects the level of neuronal activity and is essential for analyzing network dynamics, information encoding, and computational efficiency. 
The firing rate $r$ of a neuron or a layer is typically defined as the average number of spikes emitted per unit of time. For a single neuron $i$, the firing rate is calculated in Eq \ref{eq:firingRate}.

\begin{equation}
\begin{split}
r_{\text{layer}} = \frac{1}{M_{neuron}} \sum_{i=1}^{M_{neuron}} \frac{N^r_i}{T},
\end{split}
\label{eq:firingRate}
\end{equation}
where $M_{neuron}$ is the number of the neuron in a layer. $N^r_i$ describes the number of spikes neuron $i$ in a time window, while $T$ represents duration of the time window in the layer. The firing rate is expressed in Hertz, representing spikes per second. It provides insights into how effectively the network encodes input data and helps optimize sparsity and energy consumption.

%\begin{figure}[t]
%		\centering % 表示居中
%		\includegraphics[width=0.5\textwidth]{images/CM_HMID.png}
%		\caption{Confusion Matrix on NTU-Fi-HumanID Dataset. } %\label{fig:confusionMatrixs_Fi_NTU}
%\end{figure}

% \usepackage{subcaption} % 在导言区加上（不要和 subfigure 共存）

\begin{figure}[h]
\centering
\begin{subfigure}{0.24\textwidth}
    \centering
    \includegraphics[width=0.95\linewidth]{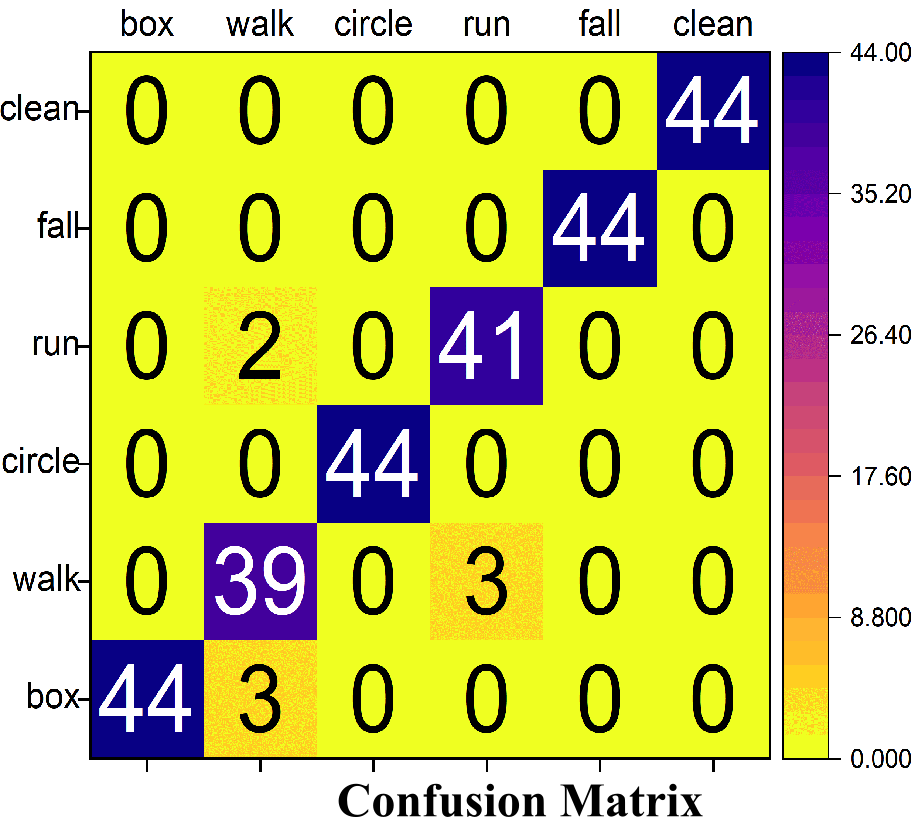}
    \caption{Confusion matrix on NTU-Fi-HAR dataset.}
    \label{fig:cm_har}
\end{subfigure}
\hfill
\begin{subfigure}{0.24\textwidth}
    \centering
    \includegraphics[width=0.95\linewidth]{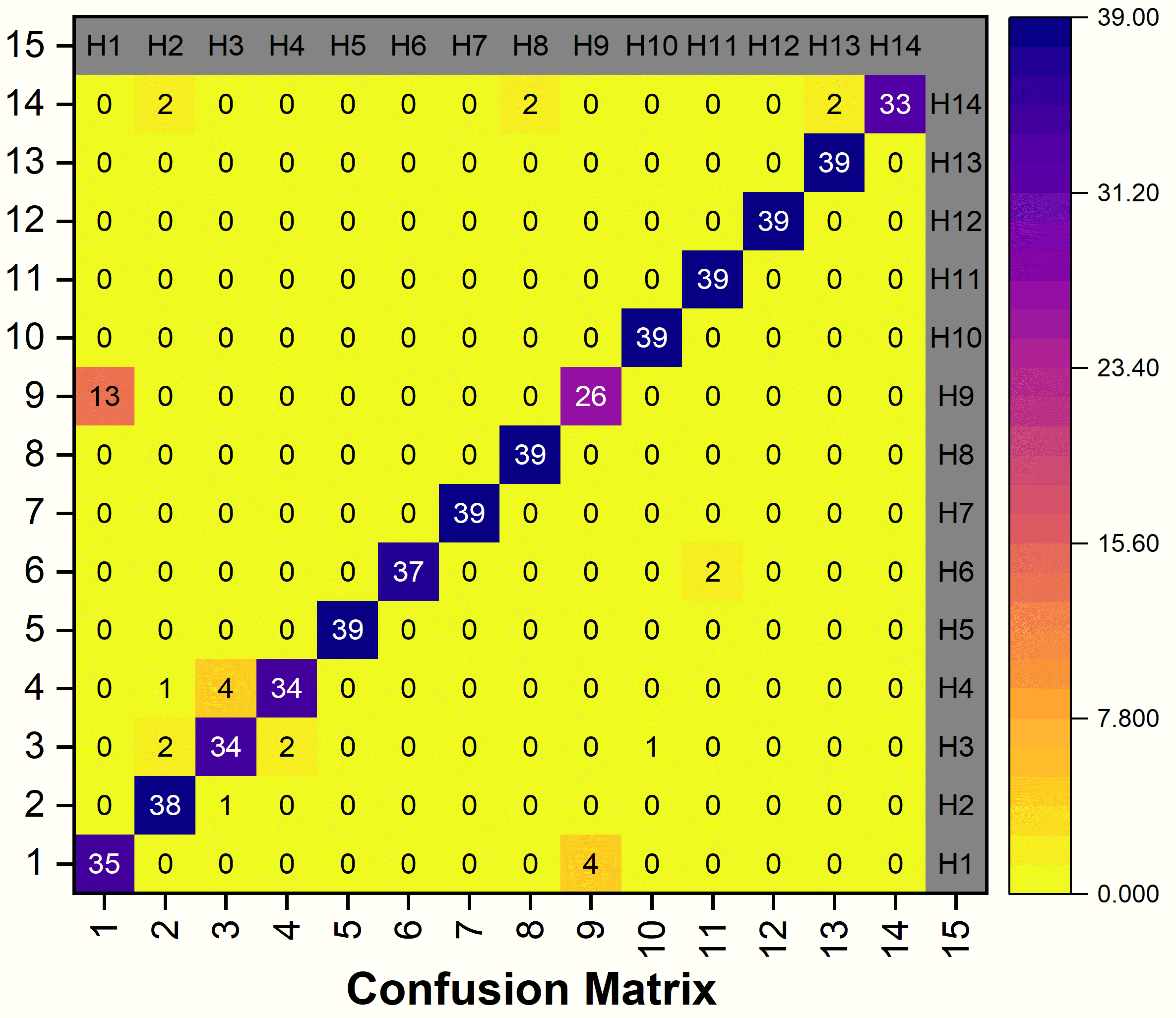}
    \caption{Confusion matrix on NTU-Fi-HumanID dataset.}
    \label{fig:cm_hmid}
\end{subfigure}
\caption{Confusion matrices on NTU-Fi-HAR and NTU-Fi-HumanID datasets.}
\label{fig:confusions}
\end{figure}

\subsection{Single Action human action recognition Comparison Experiments}

To evaluate the effectiveness of the proposed Wi-Spike model in WiFi-based human action recognition, we conducted classification experiments on three benchmark datasets: UT-HAR, NTU-Fi-HumanID, and NTU-Fi-HAR. The evaluation metrics include classification accuracy, precision, and recall, which allow us to better compare the sensing capability of Wi-Spike against existing models. In addition, we provide the confusion matrix results to further illustrate the model’s discriminative ability across different WiFi action categories.

In order to analyze the Wi-Spike model's ability to discriminate specific human action CSI signals, its confusion matrices are presented in NTU-Fi-HumanID and NTU-Fi-HAR (see Fig. \ref{fig:confusions} (a) and Fig. \ref{fig:confusions} (b)). As shown in Fig. \ref{fig:confusions} (a), the Wi-Spike model does not achieve clear separability between “walking” and “running”; compared with other categories, some misclassifications occur between these two classes. Specifically, three “running” samples were misclassified as “walking,” while two “walking” samples are predicted as “running”. Additionally, three “walking” samples are incorrectly classified as “boxing”. This confusion primarily arises from the high similarity between walking and running actions in terms of body motion, primarily differing in movement frequency. Consequently, the corresponding CSI signals are also highly similar. Although Wi-Spike demonstrates robust recognition performance, distinguishing between these similar actions remains a challenging task. As shown in Fig. \ref{fig:confusions} (b), the misclassifications are also concentrated within a limited number of categories; nevertheless, the model exhibits overall satisfactory performance.

To validate its classification results in standard WiFi-based human action recognition tasks, we report the performance of Wi-Spike on single-action classification using CSI signals. Specifically, we present key evaluation metrics on the UT-HAR, NTU-Fi-HumanID, and NTU-Fi-HAR datasets, including accuracy, F1-score, precision, and recall (see Tables \ref{tab:NTU_HAR}, \ref{tab:WiFi_humanID}, and \ref{tab:WiFi_UT}).

The results demonstrate that Wi-Spike delivers robust performance across the evaluated datasets, starting with NTU-Fi-HAR, where it achieves 96.97\% accuracy, 96.93\% F1-score, 97.06\% precision, and 96.90\% recall. This places it competitively among baselines, outperforming ViT (93.18\% accuracy) and CNN+GRU (92.05\% accuracy), while trailing the top performers like CNN (98.60\% accuracy) and LSTM (97.70\% accuracy); notably, Wi-Spike's high precision indicates strong reliability in positive identifications, with balanced recall suggesting effective coverage of true positives despite the dataset's demands. Moving to NTU-Fi-HumanID, a more complex dataset with more classes, Wi-Spike attains 95.58\% accuracy, 95.45\% F1-score, 96.14\% precision, and 95.58\% recall, surpassing ViT (78.91\% accuracy) and CNN+GRU (87.27\% accuracy) but slightly below leading models such as ResNet18 (96.63\% accuracy), LSTM (97.19\% accuracy), and CNN (97.10\% accuracy); a key feature here is that LSTM and CNN exhibit the highest overall metrics, particularly in precision exceeding 98\%, while Wi-Spike demonstrates balanced performance across all metrics, with its precision outperforming ViT and CNN+GRU, reflecting its robustness in handling increased category diversity and complex classification challenges. Finally, on UT-HAR, Wi-Spike reaches 96.51\% accuracy, 95.85\% F1-score, 96.34\% precision, and 96.98\% recall, exceeding LSTM (87.45\% accuracy), ViT (96.25\% accuracy), and CNN+GRU (96.18\% accuracy), while approaching the best baselines like ResNet18 (98.00\% accuracy) and CNN (97.61\% accuracy); standout aspects include its superior recall over most competitors, underscoring enhanced recognition of relevant actions, and overall metrics that highlight balanced performance in a standard WiFi-based recognition task. Collectively, 
these outcomes underscore the efficacy of our SNN-based method for single-action WiFi action recognition, with minor variations linked to dataset complexity, and pave the way for further analysis of its energy efficiency benefits in upcoming experiments.

\begin{table}[h]
    \centering
    \caption{The classification results on NTU-Fi-HAR dataset.}
    \label{tab:NTU_HAR}
    \begin{tabular}{ccccc}
    \hline
    Algorithm & Accuracy & F1-score & Precision    & Recall   \\ \hline
     ours     &$96.97\%$ &$96.93\%$ &$97.06\%$     & $96.90\%$      \\
     ViT      &$93.18\%$ &$93.02\%$ &$93.95\%$     & $93.08\%$     \\
     Resnet18 &$95.19\%$ &$95.20\%$ &$95.21\%$     & $95.19\%$       \\
     CNN+GRU  &$92.05\%$ &$91.92\%$ &$93.19\%$     & $92.05\%$     \\
     LSTM     &$97.70\%$ &$96.95\%$ &$98.19\%$     &    $96.96\%$  \\
     CNN     &$98.60\%$ &$98.72\%$ &$98.7\%$     &    $96.96\%$ \\
    \hline
    \end{tabular}
\end{table}

\begin{table}[h]
    \centering
    \caption{The classification results on NTU-Fi-HumanID dataset.}
    \label{tab:WiFi_humanID}
    \begin{tabular}{ccccc}
    \hline
    Algorithm & Accuracy    & F1-score   & Precision        & Recall     \\ \hline
     ours     &  $95.58\%$  &$95.45\%$   &     $96.14\%$    & $95.58\%$  \\
     ViT      &  $78.91\%$  & $78.38\%$  &     $79.34\%$    & $78.91\%$     \\
     ResNet18 &  $96.63\%$  &$96.45\%$   &     $97.26\%$    &$96.54\%$   \\
     CNN+GRU  &  $87.27\%$  &$86.40\%$   &     $88.37\%$    &  $87.07\%$  \\
     LSTM     &  $97.19\%$  & $96.74\%$  &     $98.84\%$    &  $97.13\%$  \\
     CNN     &$97.10\%$     &$96.55\%$   &$98.65\%$     &    $96.60\%$   \\
    \hline
    \end{tabular}
\end{table}

\begin{table}[h]
    \centering
    \caption{The classification results on UT-HAR dataset.}
    \label{tab:WiFi_UT}
    \begin{tabular}{cccccc}
    \hline
    Algorithm & Accuracy    & F1-score   & Precision        & Recall      \\ \hline
     ours     &  $96.51\%$  &$95.85\%$   &     $96.34\%$    & $96.98\%$   \\
     ViT      &  $96.25\%$  & $96.28\%$  &     $96.33\%$    & $96.09\%$      \\
     ResNet18 &  $98.00\%$  &$97.99\%$   &   $98.02\%$      &$97.99\%$    \\
     CNN+GRU  &  $96.18\%$  &$96.10\%$   &     $96.33\%$    &  $96.08\%$  \\
     LSTM     &  $87.45\%$  & $86.62\%$  &     $88.73\%$    &  $87.14\%$  \\
     CNN     &$97.61\%$     &$97.65\%$   &$97.15\%$     &    $96.97\%$   \\
    \hline
    \end{tabular}
\end{table}

\subsection{Experiments on Multi-Action WiFi-Based Action Recognition}

\begin{figure}
		\centering % 表示居中
		\includegraphics[width=0.5\textwidth]{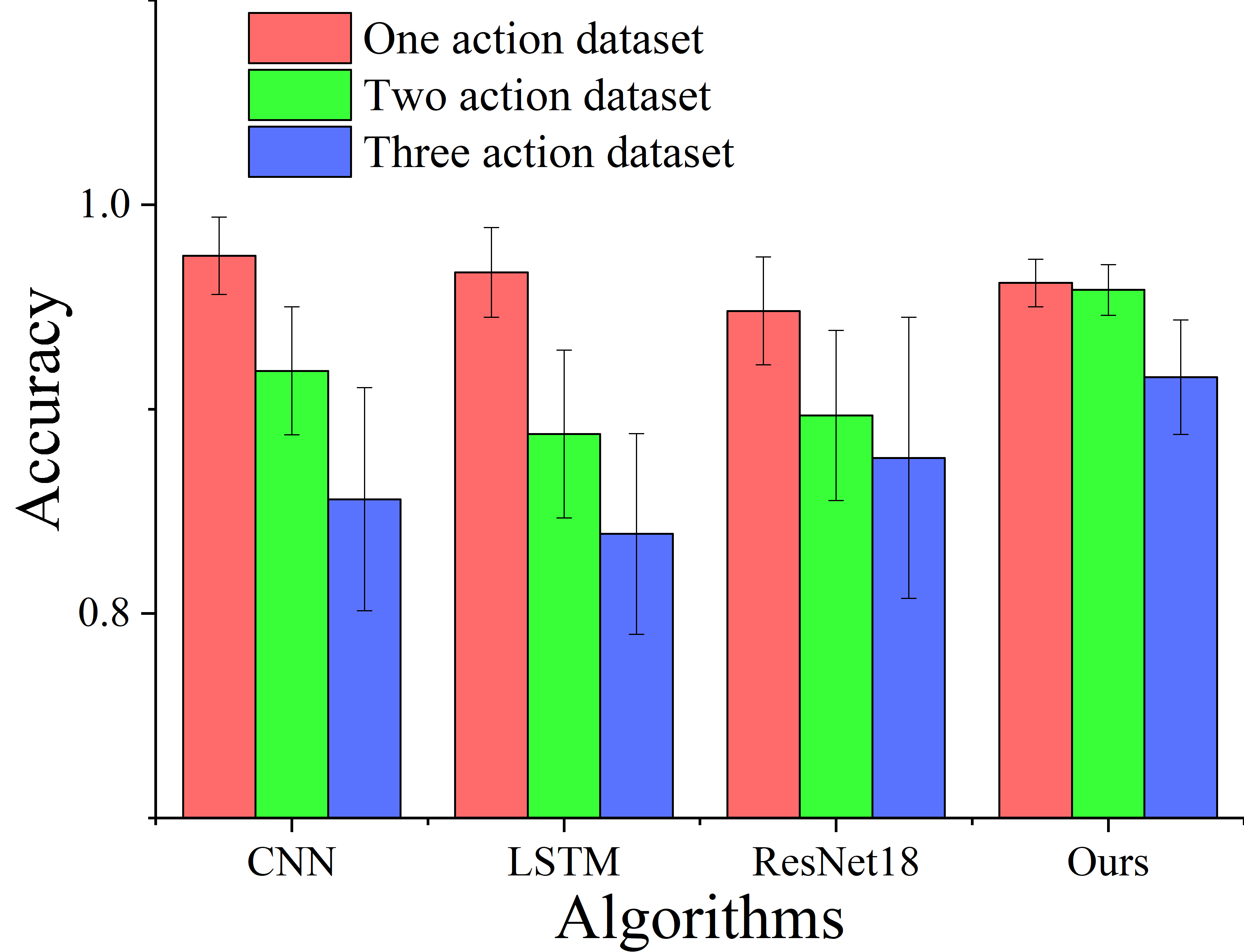}
		\caption{Comparison of Algorithm Accuracy across Different action lengths in NTU-Fi-HAR datasets. } 
        \label{fig:difActions}
\end{figure}

To evaluate the robustness of the proposed Wi-Spike model in recognizing complex activities, we conducted WiFi CSI-based multi-action human behavior recognition experiments. The primary objective of this experiment is to address the recognition of multi-action or composite activities, which are widely present in real-world scenarios. Compared with traditional simple and single-action WiFi CSI signals, multi-action WiFi signals are characterized by higher complexity, longer temporal contexts, and more diverse signal patterns. As a result, conventional neural network–based sensing models often struggle to handle such complex CSI data, leading to performance degradation. In our experiment, firstly, we evaluate the performance of the models on dual-action datasets using accuracy, F1-score, precision, and recall, and compare the results with other baseline models. Then, we further analyze how the recognition accuracy of all models changes with the increase in action length (from one action to three actions), in order to highlight the robustness of the Wi-Spike model.

In the dual-action classification experiments, we compared Wi-Spike against several baseline models, including VIT, ResNet18, CNN+GRU, LSTM, and CNN, on two benchmark datasets: NTU-Fi-HAR (double-action) and NTU-Fi-HumanID (double-action). As shown in Table \ref{tab:NTU_HAR_double} for NTU-Fi-HAR double dataset, Wi-Spike achieves the highest accuracy of $95.83\%$, outperforming the next best (CNN at $91.87\%$) by approximately $4\%$, with superior F1-score (95.79\%), precision ($96.38\%$), and recall ($95.83\%$). Similarly, Table \ref{tab:WiFiData_huamnID_double} illustrates results on NTU-Fi-HumanID double dataset, where Wi-Spike attains $92.12\%$ accuracy, surpassing ResNet18 ($90.75\%$) and LSTM ($91.50\%$), alongside balanced metrics (F1-score: $91.02\%$, precision: $93.58\%$, recall: $92.12\%$). These results underscore Wi-Spike's effectiveness in capturing intricate CSI patterns in multi-action contexts, yielding leading performance across all evaluation criteria due to its temporal attention module.

To demonstrate the robustness of the Wi-Spike model for multi-action classification tasks, Fig. \ref{fig:difActions} presents a comparison of the algorithm's accuracy on the NTU-Fi-HAR dataset under different action lengths. We conducted experiments on the NTU-Fi-HAR dataset for one action (red), two actions (green), and three actions (blue) scenarios. The experiments are repeated ten times, and the mean and variance are recorded. As shown in the bar chart, it is evident that as the number of actions increases, the accuracy of all models decreases, and the variance of the experiments also increases. However, Wi-Spike maintains the highest accuracy (e.g., consistently over 90\% in action recognition across all scenarios) with the smallest variance, indicating superior stability. This robustness stems from the model's spiking mechanism, which uses temporal spike signals to perceive changes in the time dimension of WiFi human action samples, enabling effective handling of spatiotemporal dependencies in CSI signals. Compared to baseline models such as CNN and LSTM, Wi-Spike can still maintain high recognition accuracy in the three actions scenario.

\begin{figure}
		\centering % 表示居中
		\includegraphics[width=0.5\textwidth]{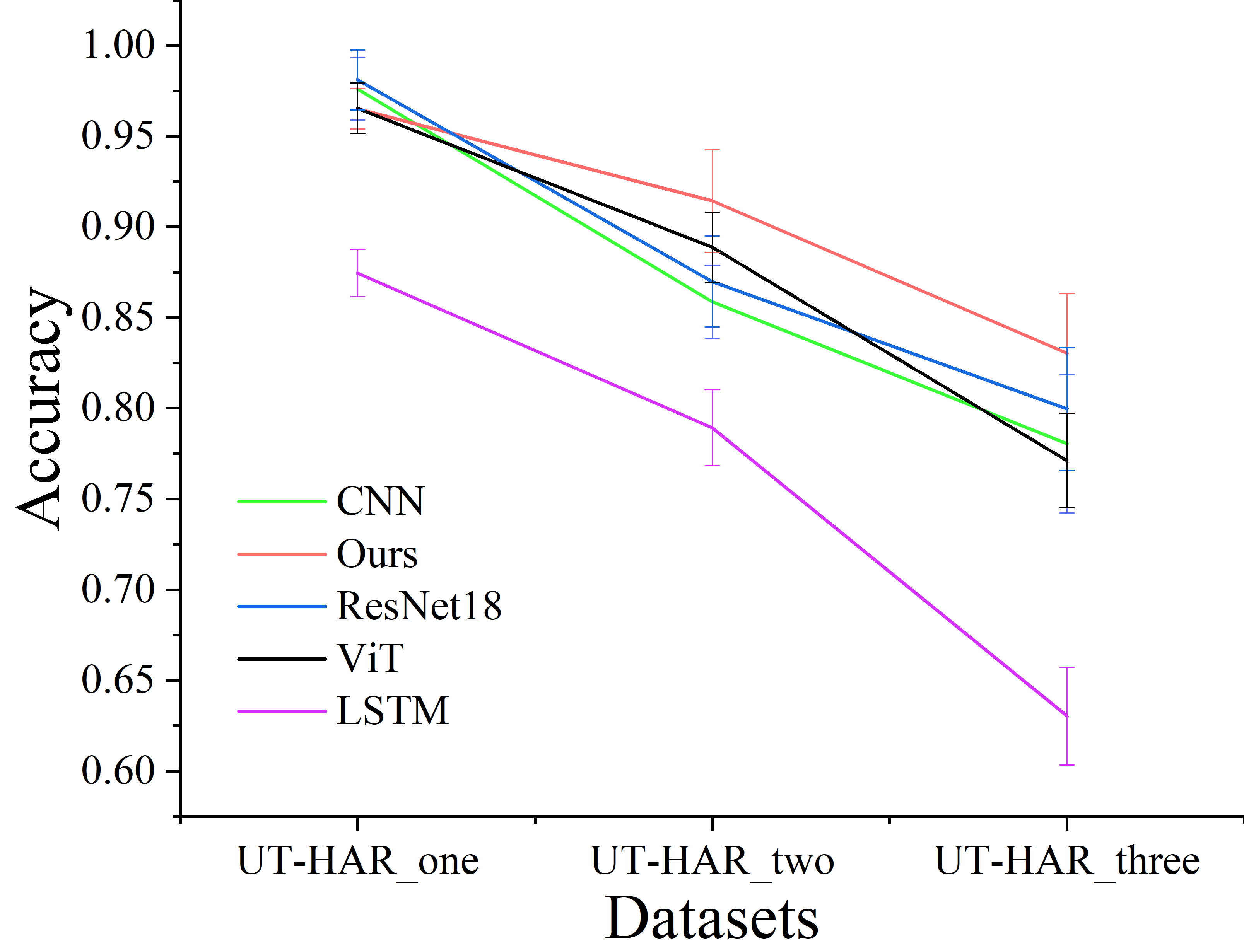}
		\caption{Comparison of classification accuracy among different models on the UT-HAR datasets. The UT-HAR\_one, UT-HAR\_two, and UT-HAR\_three datasets contain different sets of action information, allowing evaluation of model robustness under varying action configurations. } \label{fig:UT_comparsion}
\end{figure}

A similar trend is observed on the UT-HAR dataset, where Fig. \ref{fig:UT_comparsion} illustrates the line charts of Wi-Spike and baseline models (CNN, ResNet18, VIT, LSTM) under one action, two actions, and three actions scenarios. Likewise, as the number of actions increases, the recognition accuracy of all models decreases to varying degrees. However, the proposed Wi-Spike model achieves relatively high recognition accuracy even in the three action scenarios, confirming its advantage in multi-action recognition with WiFi CSI signals. Specifically, the accuracy drop of Wi-Spike (from around 95\% to 85\%) is notably smaller than that of other models, with ViT experiencing a rapid decline of more than 0.25. This indicates that Wi-Spike has stronger sensing capability for multi-action WiFi perception, even when faced with more complex CSI signals. This enhanced spatiotemporal perception ability is attributed to the unique temporal triggering mechanism of spiking neurons, which has been validated in other studies \cite{tavanaei2019deep, ghosh2009spiking}.

%\cite{liu2019wireless,ahmad2024wifi,zhang2024wifi,tan2022commodity,bocus2021uwb,li2025consense,zandi2024robofisense,abuhoureyah2024wifi,liu2023towards,ge2022contactless}

\begin{table}[h]
    \centering
    \caption{The classification experiment on NTU-Fi-HAR double dataset.}
    \label{tab:NTU_HAR_double}
    \begin{tabular}{ccccc}
    \hline
    Algorithm & Accuracy & F1-score & Precision    & Recall   \\ \hline
     ours     &$95.83\%$ &$95.79\%$ &$96.38\%$     & $95.83\%$      \\
     ViT      &$86.54\%$ &$86.40\%$ &$86.60\%$     & $86.65\%$      \\
     ResNet18 &$89.69\%$ &$86.99\%$ &$90.44\%$     & $87.50\%$       \\
     CNN+GRU  &$84.38\%$ &$76.38\%$ &$88.12\%$     & $81.06\%$      \\
     LSTM     &$88.78\%$ &$87.74\%$ &$88.84\%$     & $89.18\%$  \\
      CNN     &$91.87\%$ &$89.59\%$ &$92.94\%$     & $90.15\%$  \\
    \hline
    \end{tabular}
\end{table}

\begin{table}[h]
    \centering
    \caption{The classification experiment on NTU-Fi-HumanID double dataset.}
    \label{tab:WiFiData_huamnID_double}
    \begin{tabular}{ccccc}
    \hline
    Algorithm & Accuracy      & F1-score   & Precision     & Recall   \\ \hline
     ours     &  $92.12\%$    &$91.02\%$   &     $93.58\%$ &  $92.12\%$      \\
     ViT      &  $70.29\%$    & $70.78\%$  &     $71.34\%$ & $70.41\%$   \\
     ResNet18 &  $90.75\%$    &$88.06\%$   &     $92.82\%$ &  $89.46\%$       \\
     CNN+GRU  &  $77.50\%$    &$76.88\%$   &     $69.05\%$ &  $78.21\%$        \\
     LSTM     &  $91.50\%$    & $91.25\%$  &  $91.78\%$    &  $91.35\%$        \\
     CNN      &  $88.81\%$    &$87.58\%$   &$90.33\%$      &  $88.08\%$  \\
    \hline
    \end{tabular}
\end{table}

%5\begin{table}[h]
%    \centering
%    \caption{The experimental model on UT-HAR dataset.}
%%    \label{tab:CommunicationCode}
%%    \begin{tabular}{ccccc}
%    \hline
%    Algorithm & HumanID & HumanID\_D & HAR\_D &  Model Size  \\ \hline
%     ours  &  $96.35\%$    &$95.37\%$     &     $95.45\%$     & 481.60 kb   \\
  %   ViT  &  $93.75\%$    & $96.53\%$     &     $53.41\%$      & %3.19M\\
  %   Resnet18  &  $95.31\%$    &$98.11\%$     &     $91.67\%$  & %42.74M   \\
  %   CNN+GRU  &  $93.75\%$    &$96.72\%$     &     $77.65\%$   & %228,36kb   \\
  %   LSTM  &  $97.14\%$    & $87.18\%$     &     $86.36\%$     & %411.53kb \\
 %    CNN(current)  &  $97.14\%$    & $87.18\%$     &     $86.36\%$     %& 411.53kb \\
%    \hline
%    \end{tabular}
%\end{table}

\begin{figure}
		\centering % 表示居中
		\includegraphics[width=0.5\textwidth]{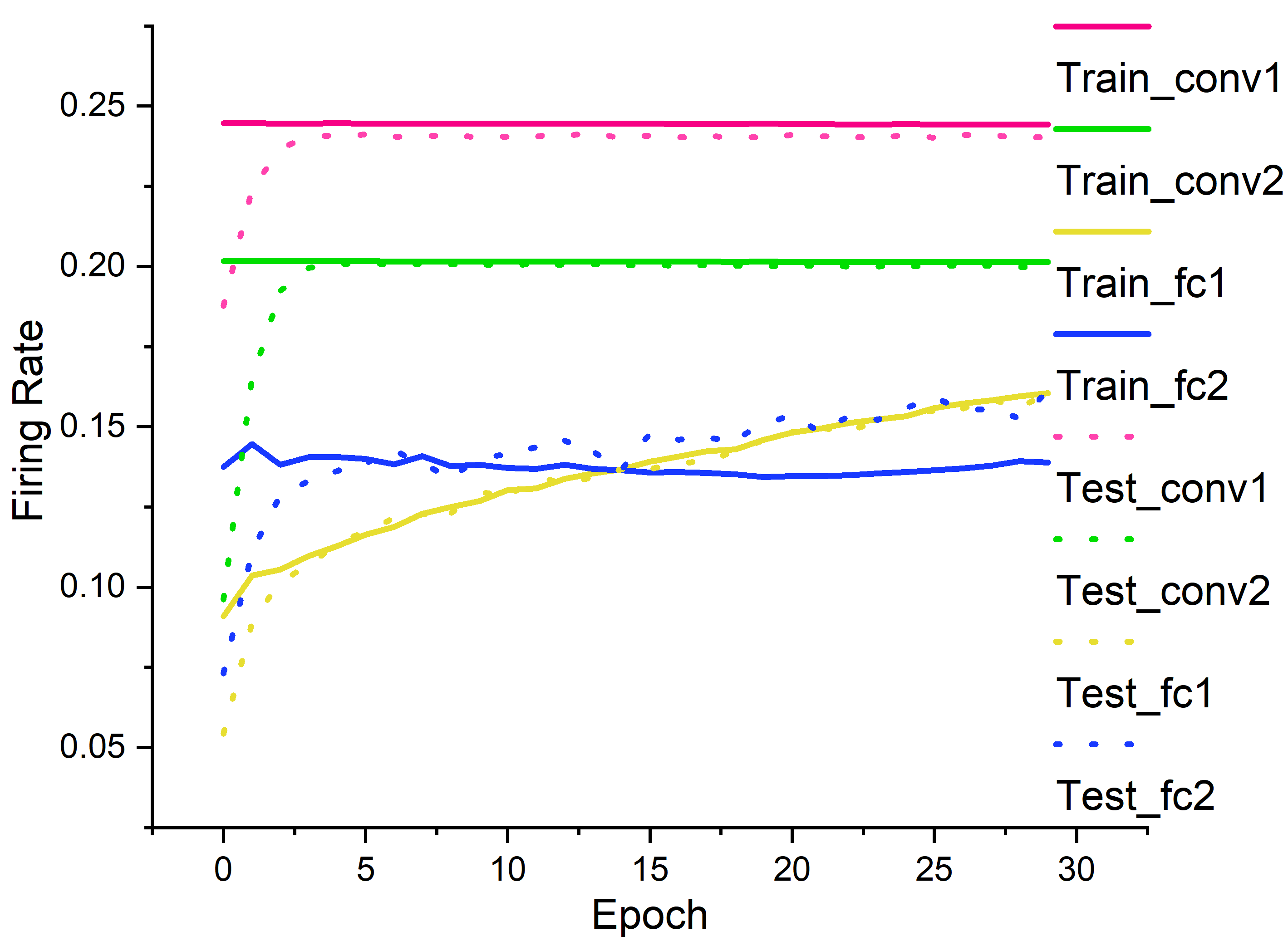}
		\caption{Firing rate of each network layer across training epochs for both training and testing phases. } \label{fig:firingRate}
\end{figure}

\subsection{Firing Rate and Model Energy Analysis in The Proposed Model}
To comprehensively evaluate the effectiveness of the proposed Wi-Spike model, we conduct experiments on firing rate dynamics and energy consumption. The firing rate analysis serves as a critical diagnostic indicator of spiking neural networks, reflecting whether the network operates within a biologically inspired and computationally stable regime. Typically, an average firing rate between 0.1 and 0.3 is considered optimal \cite{stanojevic2024high}, as it ensures sufficient neural activity for information transmission while avoiding excessive energy consumption. Meanwhile, energy analysis provides valuable insights into the trade-off between recognition accuracy and computational cost. Although Wi-Spike may not always yield the absolute highest recognition accuracy, its superior balance of accuracy and power efficiency highlights the model’s suitability for deployment in resource-constrained environments. Together, these evaluations verify both the functional stability and practical efficiency of the proposed approach.

\begin{figure}
		\centering % 表示居中
		\includegraphics[width=0.5\textwidth]{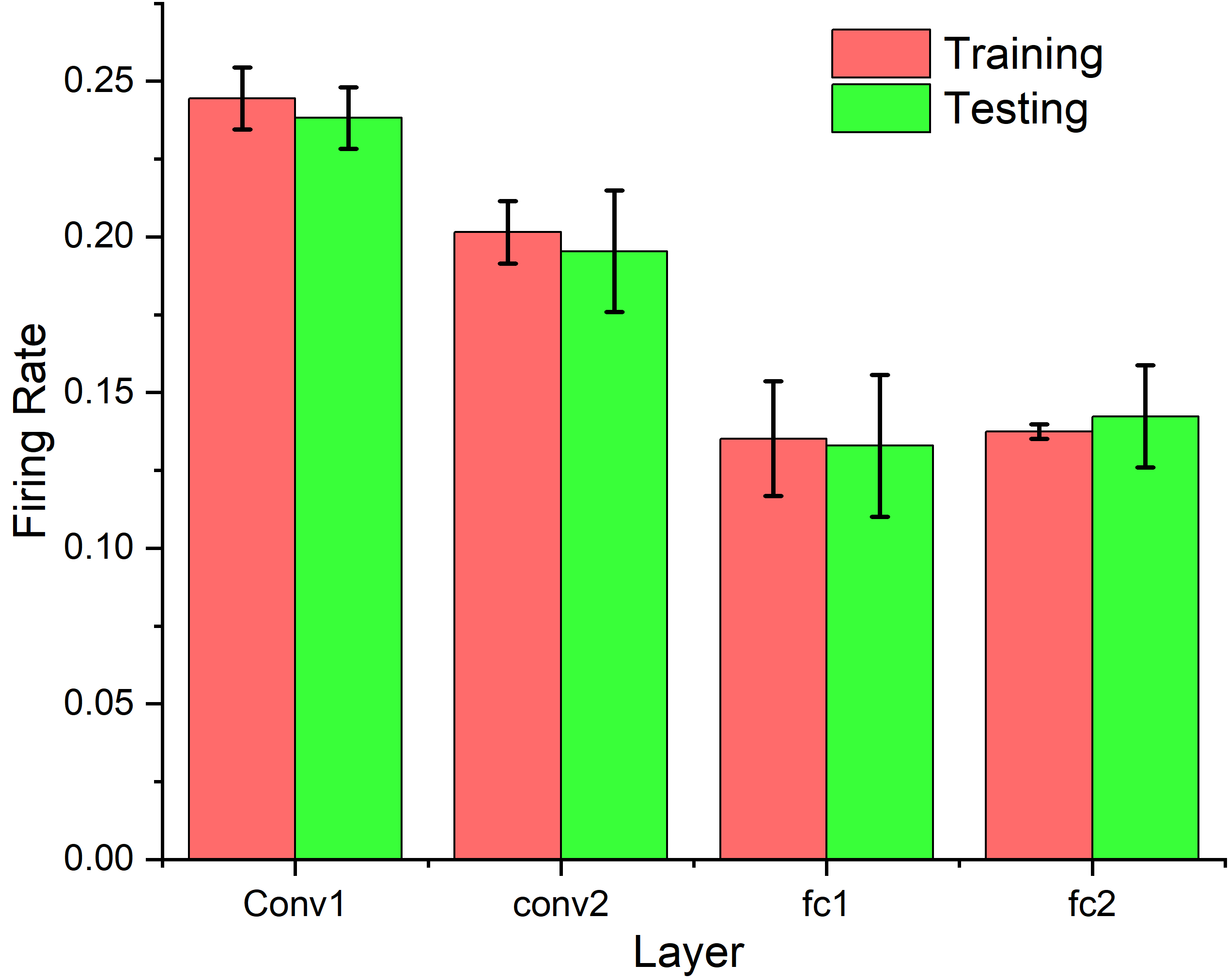}
		\caption{Mean firing rate of each network layer during training and testing. } 
        \label{fig:firingRateBar}
\end{figure}

Fig. \ref{fig:firingRate} and \ref{fig:firingRateBar} illustrate the firing rate characteristics of Wi-Spike during the training and testing processes across different network layers. As shown in Fig. \ref{fig:firingRate}, the firing rate of each layer gradually stabilizes over epochs, with values consistently converging within the 0.1–0.3 range. This indicates that Wi-Spike operates in a normal spiking regime, ensuring that the model maintains efficient neural activity without falling into under-activation or over-activation states. Furthermore, Fig. \ref{fig:firingRateBar} compares the average firing rates between training and testing phases across network layers. Notably, for each layer, the mean firing rates of training and testing sets exhibit high consistency, with closely aligned means and comparable standard deviations—evidenced by the overlapping error bars. This close alignment underscores the strong generalization capability of Wi-Spike, as it minimizes discrepancies that could indicate overfitting. Collectively, these results affirm that the model sustains stable dynamics throughout both convolutional and fully connected layers, which is crucial for reliable spiking computation and sustainable energy efficiency.

\begin{figure}
		\centering % 表示居中
		\includegraphics[width=0.5\textwidth]{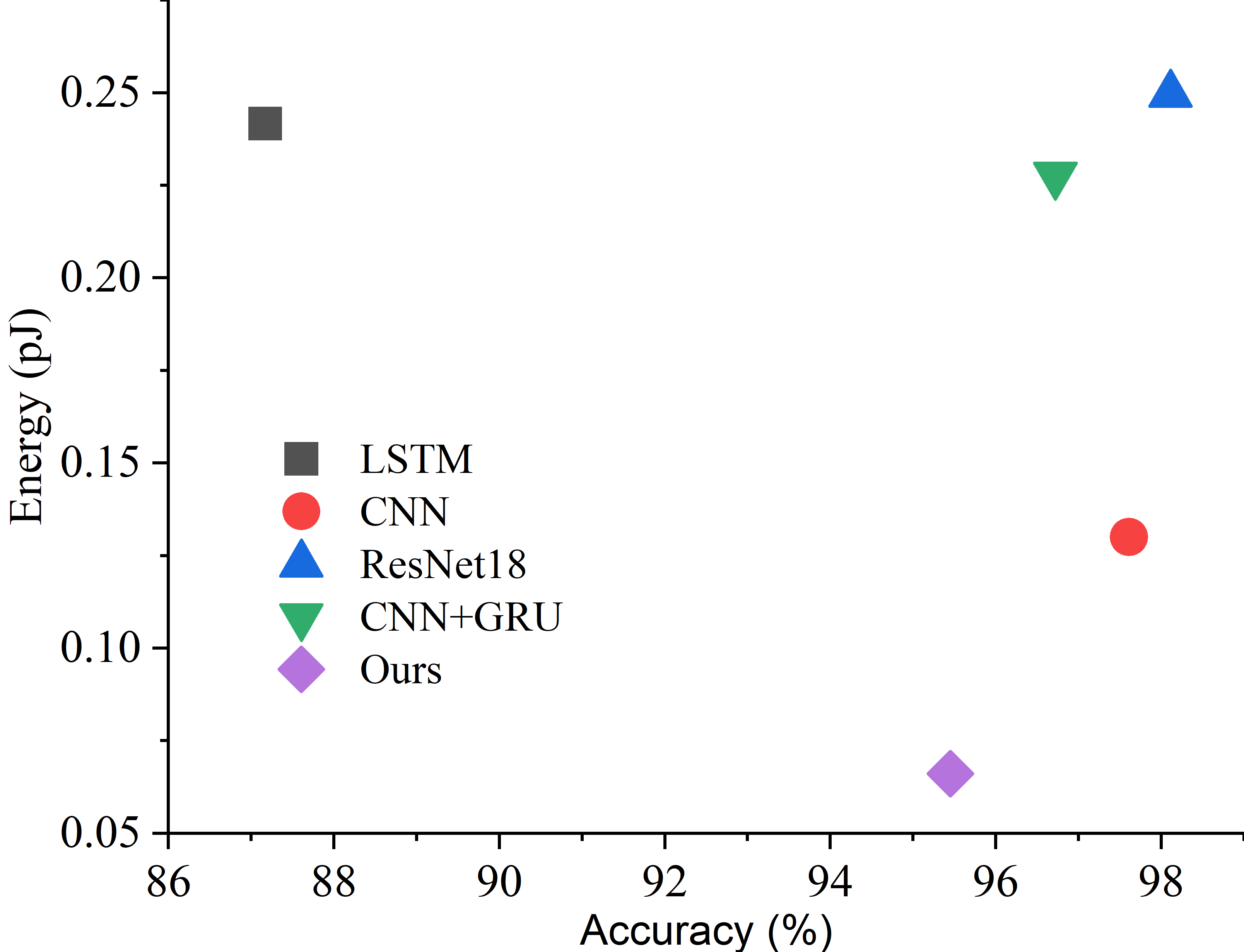}
		\caption{Comparison of model accuracy and energy consumption. } 
        \label{fig:engery}
\end{figure}

As for the energy consumption analysis of the proposed Wi-Spike model, Fig. \ref{fig:engery} presents a comparative study of recognition accuracy and energy cost on the NTU-Fi-HAR dataset across several baseline models, including CNN, ResNet18, LSTM, and CNN+GRU. The purple diamond in the figure denotes our proposed sensing model. As shown in Fig. \ref{fig:engery}, although some conventional WiFi sensing architectures achieve competitive recognition accuracy, they incur substantially higher energy costs. In contrast, Wi-Spike achieves a highly favorable balance between accuracy and energy efficiency. It delivers state-of-the-art recognition performance for WiFi-based action recognition while significantly reducing energy consumption. This advantage mainly stems from the unique event-driven mechanism of spiking neural networks, which leverage sparse spike responses to discriminate human activities from WiFi CSI signals. Consequently, Wi-Spike demonstrates an excellent trade-off between energy efficiency and recognition accuracy, highlighting its potential for energy-constrained edge applications in WiFi-based human action recognition.   

To further demonstrate the advantage of energy efficiency of Wi-Spike, we evaluate the computational efficiency and theoretical power consumption of the proposed model on the NTU-FI-HAR\_Double dataset. Unlike CNNs, which typically measure computational complexity in terms of FLOPs (floating-point operations) and MACs (multiply–accumulate operations), spiking neural networks (SNNs) are commonly evaluated using SOPs (synaptic operations). SOPs mainly focus on ACs (accumulate operations). As shown in Table \ref{tab:energyAnalysis}, under the same WiFi CSI input size ($3\times114\times500$), Wi-Spike achieve the lowest power consumption of 0.07 pJ while maintaining the highest recognition accuracy of 95.83\%. In terms of model complexity, Wi-Spike contains 14.34 M parameters and 73.51 M ACs (the theoretical amount of accumulated operations based on spike signals). According to the energy consumption principle reported in a study \cite{rathi2021diet} (based on 45 nm technology, where one AC costs 0.9 pJ and one MAC (Multiply–Accumulate operation) costs 4.6 pJ), Wi-Spike demonstrates the lowest overall energy cost for WiFi-based action recognition. Remarkably, compared with the widely adopted attention-based ViT model, Wi-Spike requires only one thirty-third of its energy to accomplish human activity sensing with WiFi signals.

% Energy consumption is based on 45nm technology, where AC cost 0.9pJ and MAC cost 4.6pJ.

% \_d
\begin{table*}[h]
    \centering
    \caption{Detailed comparison of computational efficiency and power consumption in NTU-Fi-HAR\_Double.}
    \label{tab:energyAnalysis}
    \begin{tabular}{ccccccc}
    \hline
    Algorithm & Input Size              & Energy   & Accuracy  &  MACs            &  ACs(M)    & Param    \\ \hline
     ours     &  $3\times114\times500$  &$0.07$   & $95.83\%$ &   -            & 73.51 M   & 14.34 M   \\
     ViT      &  $3\times114\times500$  &$2.33$  & $86.40\%$ &   507.1 M      &  -        & 0.83 M      \\
     ResNet18 &  $3\times114\times500$  &$0.25$  & $89.69\%$ &   54.27 M      &   -       &  11.18 M   \\
     CNN+GRU  &  $3\times114\times500$  &$0.24$  & $84.38\%$ &   53.96 M      &  -       &  0.05 M     \\
     LSTM     &  $3\times114\times500$  &$0.24$  & $88.78\%$ &   52.54 M      &  -        &  0.10 M    \\
      CNN     &  $3\times114\times500$  &$0.13$ & $91.87\%$ &   28.28 M      &  -         & 0.47 M    \\
    \hline
    \end{tabular}
\end{table*}

\subsection{Feature Map Visualization and Convergence Analysis.}

To better understand how the Wi-Spike model processes WiFi CSI signals for human action recognition, we conducted convergence analysis and feature map visualization experiments on the NTU-Fi-HAR datasets. 

\begin{figure}
		\centering % 表示居中
		\includegraphics[width=0.5\textwidth]{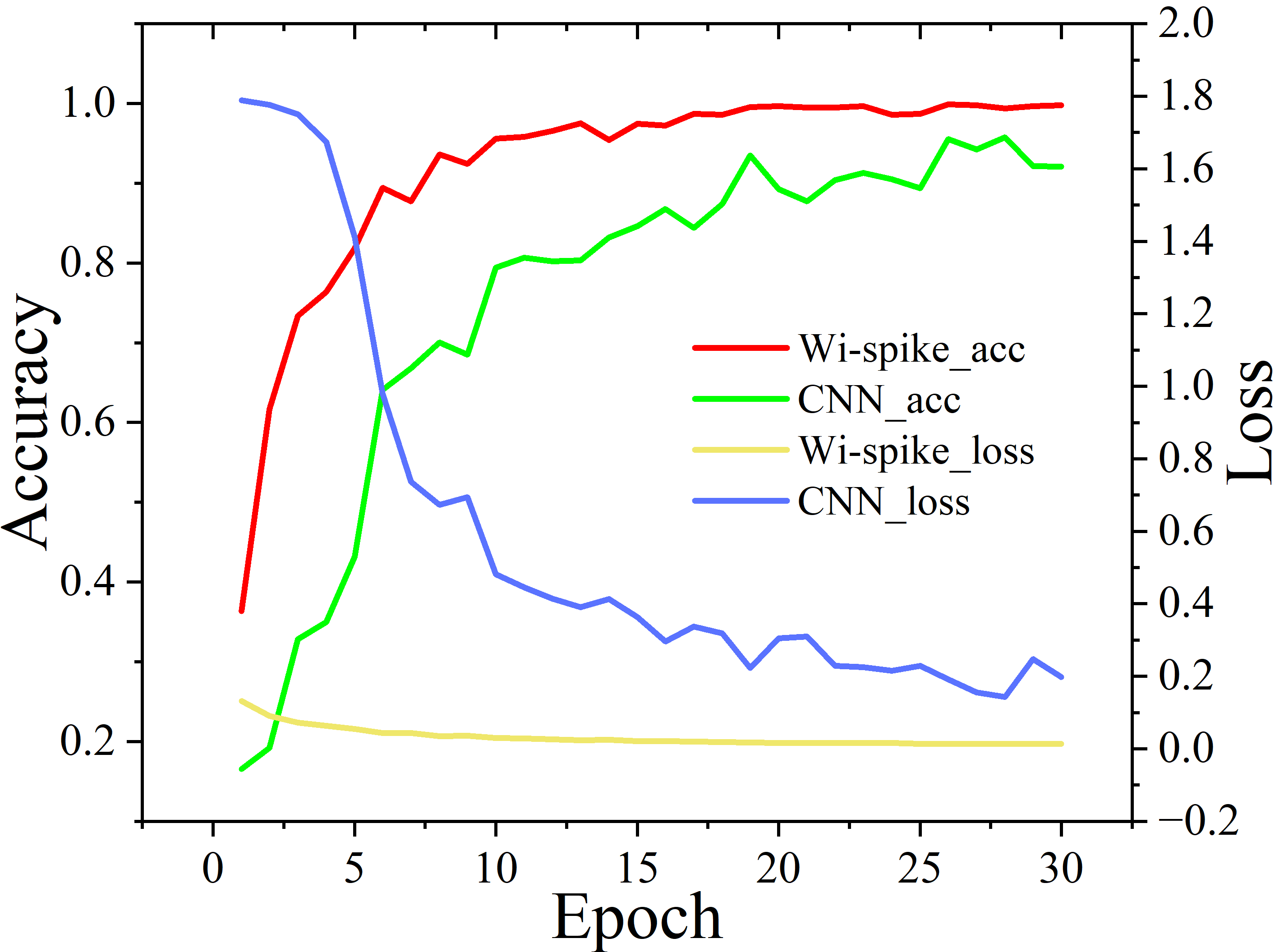}
		\caption{Comparison of accuracy and loss changes during neural networks training. } \label{fig:feature_maps}
\end{figure}

For conververgence analysis, a standard CNN model is selected for comparison. We present the accuracy and loss curves of both stability. As shown in Fig. \ref{fig:feature_maps}, Wi-Spike (red and blue lines) demonstrates a faster accuracy increase and a quicker loss reduction compared to CNN (green and yellow lines), ultimately converging at a higher performance level. This result illustrates the fundamental advantage of spiking neural networks (SNNs) in WiFi HAR task, namely their ability to achieve more efficient learning through event-driven processing, which enables Wi-Spike to realize superior training dynamics over conventional CNNs.

\begin{figure*}[htbp]
\centering

% 第一行
\begin{subfigure}{0.48\textwidth}
    \captionsetup{skip=6pt}
    \centering
    \includegraphics[width=\linewidth]{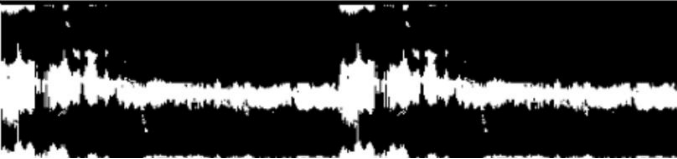}
    \caption{Wi-Spike Feature Map 1}
\end{subfigure}
\hfill
\begin{subfigure}{0.5\textwidth}
    \captionsetup{skip=6pt}
    \centering
    \includegraphics[width=\linewidth]{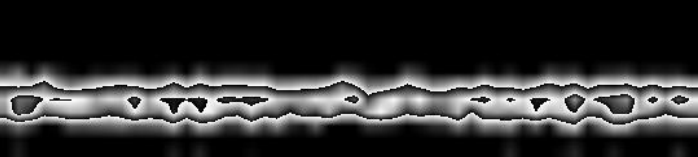}
    \caption{CNN Feature Map 1}
\end{subfigure}

\vspace{0.8em} % 两行之间的垂直间距

% 第二行
\begin{subfigure}{0.47\textwidth}
    \captionsetup{skip=6pt}
    \centering
    \includegraphics[width=\linewidth]{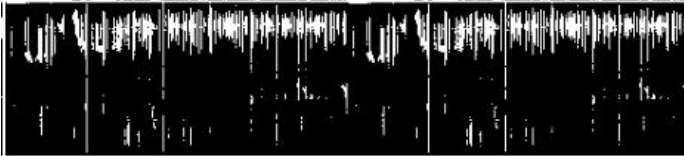}
    \caption{Wi-Spike Feature Map 2}
\end{subfigure}
\hfill
\begin{subfigure}{0.505\textwidth}
    \captionsetup{skip=6pt}
    \centering
    \includegraphics[width=\linewidth]{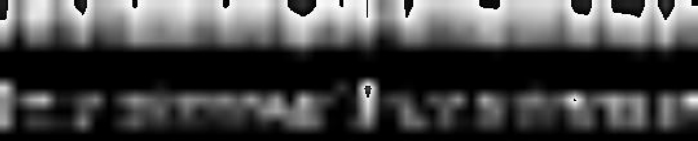}
    \caption{CNN Feature Map 2}
\end{subfigure}

\vspace{0.8em}

% 第三行
\begin{subfigure}{0.48\textwidth}
    \captionsetup{skip=6pt}
    \centering
    \includegraphics[width=\linewidth]{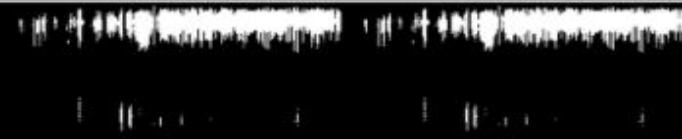}
    \caption{Wi-Spike Feature Map 3}
\end{subfigure}
\hfill
\begin{subfigure}{0.5\textwidth}
    \captionsetup{skip=6pt}
    \centering
    \includegraphics[width=\linewidth]{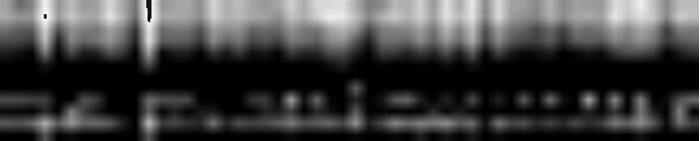}
    \caption{CNN Feature Map 3}
\end{subfigure}

\caption{Comparison of feature maps after first layer between Wi-spike (left) and traditional CNN (right)}
\label{fig:grid}
\end{figure*}

For feature visualization, Fig. \ref{fig:grid} further compares the feature maps of Wi-Spike and CNN using a dual-action WiFi CSI sample from the NTU-Fi-HAR double dataset. The SNN-based visualization (left) reveals clearer signal envelopes and finer details of multiple human activities, with well-defined patterns observed in Fig. \ref{fig:grid} (a), (c), and (e). In contrast, the CNN visualization (right) appears blurrier and contains fewer details in Fig. \ref{fig:grid} (b), (d), and (f). These results suggest that Wi-Spike captures subtle variations in multi-action human WiFi CSI signals more effectively, leading to improved recognition capability beyond CNN. This advantage can be attributed to the spike-based encoding mechanism, which preserves the temporal dynamics of CSI data, making SNNs inherently well-suited for complex time-series sensing tasks, such as WiFi-based multi-action human action recognition.

%\begin{table*}[h]
%    \centering
%    \caption{The experimental communication code in three flight test.}
%    \label{tab:comparison}
%    \begin{tabular}{cccccc}
%    \hline
%    Algorithm & Input Size    & Energy       &  HAR            &  Ops                & Param  \\ \hline
%     ours     &  $3\times114\times500$    &$0.066$     &     $95.45\%$   & 73.51 M   & \\
%     ViT      &  $3\times114\times500$    & $2.332$    &     $53.41\%$   & 507.1 M   &  \\
%     Resnet18 &  $3\times114\times500$    &$0.2496$     &     $91.67\%$   & 54.27 M  & \\
%     CNN+GRU  &  $3\times114\times500$    &$0.2278$     &     $77.65\%$   & 49.54 M  & \\
%     LSTM     &  $3\times114\times500$    & $0.2416$    &     $86.36\%$   & 52.54 M  & \\
%      LeNet   &  $3\times114\times500$    & $0.13008$    &     $89.02\%$   & 28.28 M &  \\
%    \hline
%    \end{tabular}
%\end{table*}

\section{Conclusion}\label{6}
In this work, we address the problem of WiFi-based human multi-action recognition and propose a novel low-power model, Wi-Spike. Similar to conventional neural network–based WiFi CSI classification models, Wi-Spike performs human action classification by recognizing raw WiFi CSI samples. Unlike previous neural network approaches, however, Wi-Spike is designed upon an event-driven neural paradigm. Specifically, the raw WiFi CSI signals are modeled as spike sequences with temporal dimension $T$, and activity patterns are classified based on these spiking events.  In our architecture, Wi-Spike employs a spiking convolutional layer to extract features from spike-encoded CSI signals. Then, a temporal attention layer is further introduced to guide the model in capturing the most discriminative features for action recognition. Finally, a combination of spiking fully connected layers and a voting layer is used to encode and classify the CSI signals. Our experimental results demonstrate that Wi-Spike achieves highly competitive classification performance ($96.97\%$ in NTU-Fi-HAR, $95.58\%$ in NTU-FI-HumanID, and $96.51\%$ in UT-HAR) under single-action recognition tasks while maintaining minimal energy consumption. Moreover, in multi-action recognition scenarios, Wi-Spike exhibits remarkable robustness and state-of-the-art performance. Looking ahead, we plan to further optimize Wi-Spike from the perspectives of energy efficiency and recognition effectiveness, aiming to design even more lightweight WiFi sensing models tailored for stringent edge-sensing environments.

% if have a single appendix:
%\appendix[Proof of the Zonklar Equations]
% or
%\appendix  % for no appendix heading
% do not use \section anymore after \appendix, only \section*
% is possibly needed

% use appendices with more than one appendix
% then use \section to start each appendix
% you must declare a \section before using any
% \subsection or using \label (\appendices by itself
% starts a section numbered zero.)
%

\appendices

%\section*{Acknowledgment}

%The authors would like to thank...

% Can use something like this to put references on a page
% by themselves when using endfloat and the captionsoff option.
\ifCLASSOPTIONcaptionsoff
  \newpage
\fi

\bibliographystyle{IEEEtran}
\bibliography{IEEEexample}

\end{document}